\documentclass[journal]{IEEEtran}

\IEEEoverridecommandlockouts
\usepackage{pifont}
\usepackage{amsmath} 
\usepackage{amssymb}  
\usepackage{booktabs}
\usepackage[dvips]{graphicx}
\usepackage{subfigure}
\usepackage{multirow}
\usepackage{caption}
\usepackage{amsfonts}
\usepackage{tabularx}
\usepackage{amsfonts}
\usepackage{hyperref}
\usepackage{algorithm,algorithmic}
\usepackage{diagbox}
\usepackage{bm}
\usepackage{breakurl}
\usepackage{wrapfig}
\usepackage{enumerate}
\usepackage{verbatim}
\usepackage{float}
\usepackage{siunitx}
\usepackage{mdframed}
\usepackage{adjustbox}
\usepackage{cleveref}
\usepackage{authblk}
\usepackage{color}
\newcommand{\cmark}{\ding{51}}%
\newcommand{\xmark}{\ding{55}}%
\usepackage{xurl}
\usepackage{pifont}

\begin{document}

\title{Robust Active Visual Perching with Quadrotors\\ on Inclined Surfaces}
\author{Jeffrey Mao$^{1}$, Stephen Nogar$^{2}$,  Christopher Kroninger$^{2}$,  and Giuseppe Loianno$^{1}$

\thanks{$^{1}$The authors are with the New York University, Tandon School of Engineering, Brooklyn, NY 11201, USA
        {\tt\footnotesize email: \{jm7752, loiannog\}@nyu.edu}.\\
        Corresponding Author is Jeffrey Mao. ORCID ID: 0000-0001-8729-7369.\\
        $^{2}$The authors are with the DEVCOM Army Research Laboratory, 2800 Powder Mill Road, Adelphi, MD 20783, USA.
            {\tt\footnotesize email: \{stephen.m.nogar, christopher.m.kroninger\}.civ@army.mil.}}
\thanks{This work was supported by the DEVCOM ARL grant DCIST CRA W911NF-17-2-0181, NSF CAREER Award 2145277, DARPA YFA Grant D22AP00156-00, Qualcomm Research, Nokia, and NYU Wireless.}}
\markboth{IEEE Transactions on Robotics, 2022}%
{Shell \MakeLowercase{\textit{et al.}}: Bare Demo of IEEEtran.cls for IEEE Journals}

\maketitle

\begin{abstract}
Autonomous Micro Aerial Vehicles are deployed for a variety of tasks including surveillance and monitoring. Perching and staring allow the vehicle to monitor targets without flying, saving battery power and increasing the overall mission time without the need to frequently replace batteries. This paper addresses the Active Visual Perching (AVP) control problem to autonomously perch on inclined surfaces up to $90^\circ$. Our approach generates dynamically feasible trajectories to navigate and perch on a desired target location while taking into account actuator and Field of View (FoV) constraints. By replanning in mid-flight, we take advantage of more accurate target localization increasing the perching maneuver's robustness to target localization or control errors. We leverage the Karush-Kuhn-Tucker (KKT) conditions to identify the compatibility between planning objectives and the visual sensing constraint during the planned maneuver. Furthermore, we experimentally identify the corresponding boundary conditions that maximize the spatio-temporal target visibility during the perching maneuver. The proposed approach works on-board in real-time with significant computational constraints relying exclusively on cameras and an Inertial Measurement Unit (IMU).
Experimental results validate the proposed approach and show a higher success rate as well as increased target interception precision and accuracy compared to a one-shot planning approach, while still retaining aggressive capabilities with flight envelopes that include large displacements from the hover position on inclined surfaces up to 90$^\circ$, angular speeds up to 750~deg/s, and accelerations up to 10~m/s$^2$.
\end{abstract}


\IEEEpeerreviewmaketitle

\section*{Supplementary material}
\noindent \textbf{Video:}
\url{https://www.youtube.com/watch?v=ZOF2q31Asf4}
\noindent\textbf{Code:} \url{https://github.com/arplaboratory/TrajPlanARPL}

\section{Introduction}\label{sec:intro}
\begin{figure}[t]
    \centering
    \includegraphics[width=\columnwidth]{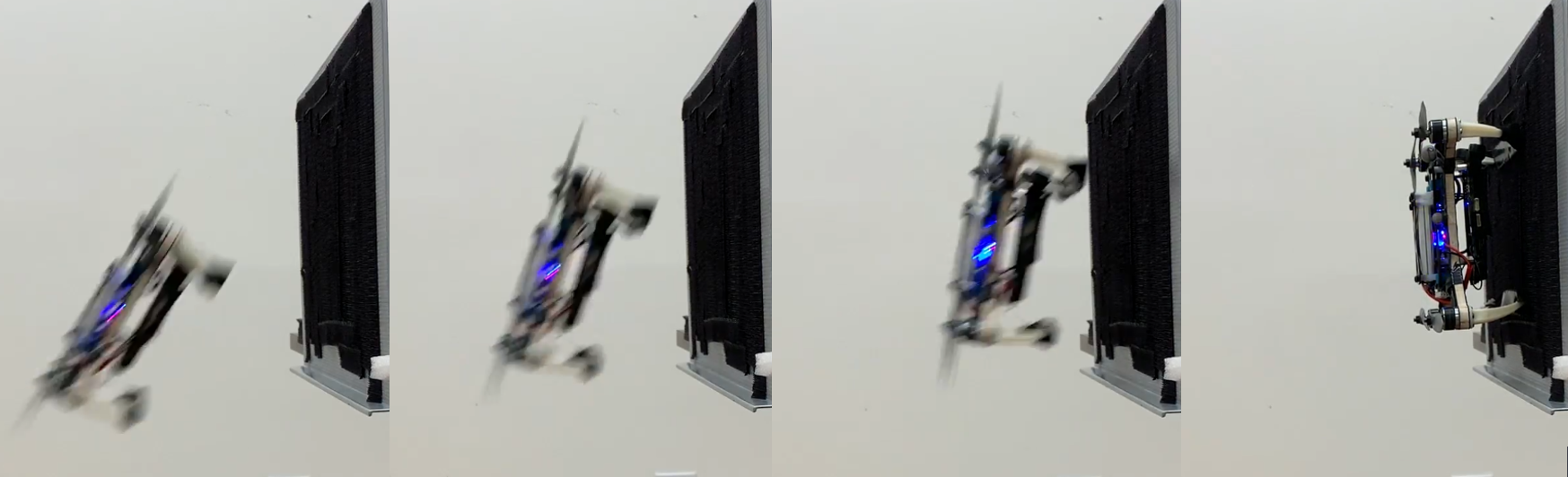}
    \caption{Aggressive visual perching sequence maneuver for a $90^\circ$ inclined surface~\cite{MAO_visaul_perch}.}
    \label{fig:90_perchingimage}
\end{figure}

\IEEEPARstart{M}{icro} Aerial Vehicles (MAVs) such as quadrotors have great speed and maneuverability while being able to hover in place. This makes them ideal for exploration and surveillance. However, MAVs are limited by low flight time in the $20-30$ minutes range. By perching on a surface, a quadrotor can extend its mission time and save power while still monitoring one or multiple targets. This motivates the need for autonomous perching solutions demonstrated in Fig.~(\ref{fig:90_perchingimage}). Inclined flat planar surfaces like walls and rooftops are plentiful especially in urban environments. By focusing on this avenue, we aim to greatly reduce the energy consumption for multiple types of missions. Several challenges complicate the perching maneuver execution. The quadrotor is an underactuated system where both orientation and acceleration of the vehicle are dependent on each other. In addition, during perching maneuvers the vehicle requires large excursions from the hover position.
In this paper, we propose Active Visual Perching (AVP) designed for robust perching with quadrotors on inclined surfaces. Our previous work~\cite{MAO_visaul_perch} addressed the state estimation, control, and planning problems for aggressive perching generating dynamically and physically feasible trajectories on planar surfaces up to $90^\circ$ solely using on-board sensing and computation perching algorithm with a one-shot planning approach. Conversely, in this work AVP involves planning new perching trajectories based on the quadrotor's subsequent target localization during the perching maneuver in real-time while concurrently maximizing its spatio-temporal visibility. This reduces the noise detection effects or partial visibility of the target by consecutively re-planning trajectories in an active fashion as well as reducing the estimation and control errors induced by the aggressive motions. Furthermore, this compensates for the use of low-resolution images and a low-cost IMU to localize the vehicle and the target. The sum of our replanned trajectories naturally favors trajectories with reduced spatial distance compared to a one-shot planning approach which further facilitates the target's interception.  

This paper presents multiple contributions. First, we present the perception, control, and planning approaches for active agile autonomous quadrotor perching on planar surfaces up to $90^\circ$ inclinations. The proposed approach is  robust to sensor and control noises as well as  to target localization errors by replanning trajectories in mid-flight. The robot exploits consecutive and more accurate target localization during the perching approach maneuver thus providing robustness compared to a one-shot planning approach presented in our previous work~\cite{MAO_visaul_perch}. Second, our strategy enforces the target visibility within the camera Field of View (FoV) to ensure the landing pad remains in sight. We exploit Karush–Kuhn–Tucker (KKT) conditions to verify the compatibility between the FoV constraint, the perching maneuver objectives, and additional actuator and sensing constraints to ensure the planned trajectory is both dynamically and physically feasible. Third, we experimentally identify the boundary conditions that can further help the FoV constraint to maximize the spatio-temporal target visibility during the maneuver. Finally, our solution is lightweight and run solely on-board limited computational unit without any external sensors or computation at 30 \si{Hz} until the target is intercepted. Multiple experimental results demonstrate that the proposed approach is able to achieve more consistent and accurate target interception for planar targets up to $90^\circ$ inclinations compared to the one-shot planning approach presented in~\cite{MAO_visaul_perch}. 

\section{Related Works}\label{sec:RW}
Perching on a vertical surface is a challenging problem because quadrotors are nonholonomic and underactuated systems where orientation depends on linear acceleration. As a result, the perching maneuver on steep inclined surfaces such as $90^\circ$ requires high angular momentum which induces target loss and camera motion blur making the localization, control, and planning problem extremely challenging. This has led to two main research avenues. The first one focuses on planning and control problems to generate feasible aggressive maneuvers. Multiple works use external sensors to localize the quadrotor and target. The second addresses the mechanical design of proposing novel mechanisms or attachments to facilitate perching on specific structures or surfaces without aggressive maneuvers.

Related to planning, \cite{Roypoly} details how to employ polynomial splines to formulate a trajectory on the flat outputs or position in the inertial frame and yaw. Further optimizations using different polynomial splines and additional costs have been proposed for solving time optimal trajectories by~\cite{wang2020alternating, FeiGao_opt_time, burke2020generating} however these works purely focus on the planning problem for aerial navigation without studying the appropriate planning constraints to resolve the perching problem. More complex planners~\cite{Watterson_SE3, liu2017searchbased} propose approaches on SE(3) that generate feasible trajectories on both rotation and translation. However, these works require substantially more computation than planning on the flat outputs requiring non-convex optimization and nonlinear constraints making them unsuitable for on-board computation on robots with limited computational units. Furthermore,  works such as~\cite{agrawal2021continuoustime, CPC_Philipp, Min_energy_traj} plan trajectories on the full robot state, but are even more computationally expensive than ~\cite{Watterson_SE3, liu2017searchbased}. Similarly for planners considering the full robot state, \cite{PowerlinePerch} tackles the perching problem specifically for power lines as opposed to planar surfaces and considers the full robot state as well as field of view and perching constraints. These approaches do not represent the trajectory based on basis functions but instead with discrete points constrained by system dynamics which are normally solved with a specialized solver nonlinear such as ACADOS \cite{acados}. Consequentially, this approach is more computationally expensive taking $400$-$1500~\si{ms}$ to solve compared to  methods based on flat outputs which generally take less than $100~\si{ms}$. Our current work is similar to ~\cite{MAO_visaul_perch,Thomas_perch_inclined, loianno_est} parameterizing the trajectory on the flat outputs and formulating the perching constraint as a linear constraints solvable with convex optimization. However, we also apply an active vision feedback planning approach to refine our trajectory and achieve more consistent interception with our $90^\circ$ inclined surfaces

Some specialized controllers perform similar aggressive maneuvers such as multiple flips in~\cite{Chen_acrobat, Lupashin_flip}. These controllers tend to be fairly narrow and designed to solely execute one maneuver such as multiple flips. They tend to not generalize very well if a different maneuver is required such as perching at a different incline and require motion capture systems. Conversely,~\cite{Kaufmann_acrobat, habas2022inverted} the authors performed aggressive motions such as quadrotor flips using solely on-board sensors, but this is based on a deep learning approach which is still very computationally expensive to run on-board small-scale robots and does not present guarantees. Other works focus on a much simpler target interception while relying purely on on-board camera. Visual Servoing approaches ~\cite{Thomas_visual_servoing,Zhang_optimal_traj} have shown controllers capable of landing on targets through purely visual feedback, but these methods are highly dependent on objects' shapes and require the object to continuously be in the FoV to preserve the control stability. Therefore, their maneuvers are not aggressive and do not present large excursions from the hover position as in our proposed case. 

 Our past~\cite{MAO_visaul_perch} and current work ensure dynamic and physical feasibility in the planned perching trajectory and executes them without external sensors, the proposed approach performs repeated target localization (using an Apriltag~\cite{olson2011tags}) and planning to actively refine the robot's motion during the maneuver to increase the perching accuracy at longer distances and robustness to both control and localization errors. Relevant to this work are other papers that consider the mid-flight replanning problem~\cite{UsenkoSPC17, RAPTOR, Penin2018, Falanga_vision_perching, ji2022real}.
However, the approaches proposed in~\cite{UsenkoSPC17, RAPTOR, Penin2018} are specifically designed for navigation and obstacle avoidance.  ~\cite{Penin2018} adds target tracking to the problem however that work specializes in nonaggressive trajectories tracking targets moving at most $1~\text{m/s}$. \cite{Falanga_vision_perching} only considers perching on targets with $0^\circ$ inclines. None of the above planners address the challenges of planning maneuvers for aggressive perching. Finally, while \cite{ji2022real} does employ a replanning framework targeted for aggressive maneuvers such as perching it uses external sensors to localize the target and does not consider the perception constraints required for maintaining the target in their drone's FoV. In comparison, we propose a more generalizable FoV constraint and a different form of anticipation to reduce errors induced by the replanning computation phase. We also employ a global bound checking algorithm to make the approach computationally efficient for multiple nonlinear constraints such as maximum thrust. Iteratively adding time and performing global bound checks provides in our case a more efficient and accurate bound solver as opposed to the process of linearizing various nonlinear actuator constraints and enforcing them at discrete time intervals across the trajectory.  

There are also a variety of perching mechanism designs that aim to simplify the perching problem on a wide range of surfaces such as claws~\cite{Chi2014,  AGRASP, roderick2021bird} for cylindrical objects, dry adhesives~\cite{Kalantari2015,Hawkes2013,Daler2013}, suction gripper mechanisms~\cite{Kessens2016}, arms attached to the quadrotor \cite{Backus_gripper_arm, wopereis_arm}, electroadhesive material~\cite{Park_ceiling}, or other passive mechanisms ~\cite{Pope_SCAMP}. Employing customized mechanisms can ease the perching problem on specific surfaces sometimes at the price of a mass increase of the vehicle that can range from $10\%$ in ~\cite{Backus_gripper_arm} to $40\%$ in~\cite{Pope_SCAMP}. Our proposed planning strategy is general and can complement any customized mechanism by imposing additional boundary constraints to robustify the perching maneuver. In this work, we employ a VELCRO\textsuperscript{\textregistered} mechanism due to its simplicity and low cost. Finally, fixed-wing solutions for vertical perching have been proposed~\cite{Lussier2011, Moore_2014,Mehanovic_fixed_wing}. Fixed-wing aircraft gain greater flight endurance, but are bulkier and have reduced maneuverability compared to quadrotors. Since quadrotors have reduced flight time compared to fixed-wing solutions, we are further motivated in our research to exploit their maneuverability to land on ideal perching locations and extend their corresponding mission time.

\section{Preliminaries}\label{sec:Methods}

\begin{figure}[!t]
    \centering
    \includegraphics[width=\linewidth]{
    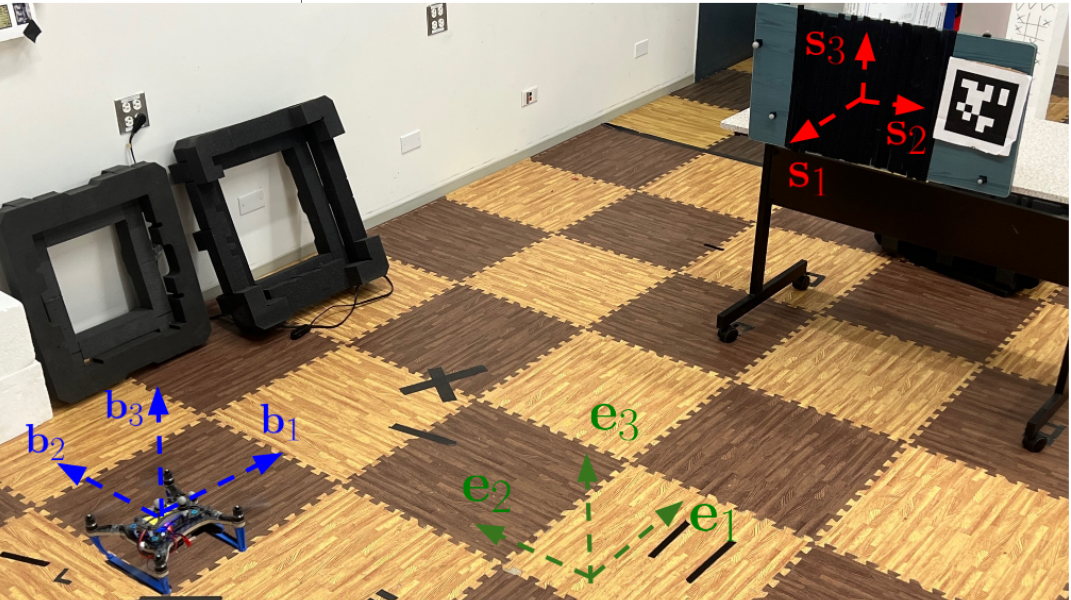}
    \caption{Visualization of the quadrotor, inertial, and target frame.}
    \label{fig:Perching Setup}
        \vspace{-10pt}
\end{figure}

\begin{figure*}[!t]
    \centering
    \includegraphics[width=\textwidth]{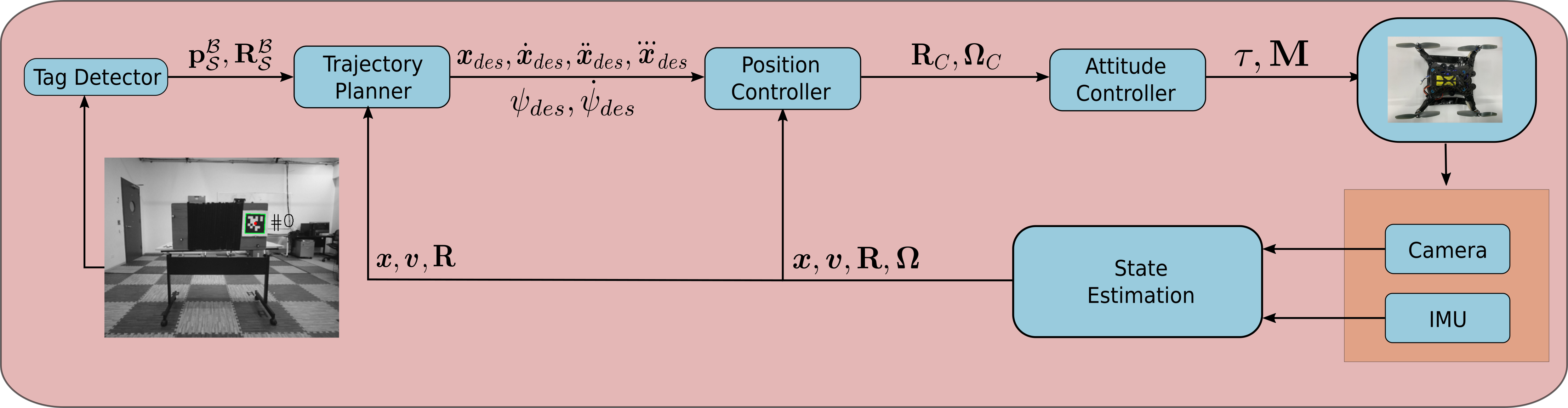}
    \caption{System architecture for the perching task.}
    \label{fig:blockdiagram}
    \vspace{-10pt}
\end{figure*}

The inertial frame $\mathcal{I}$ is defined by the three axes $\begin{bmatrix} \mathbf{e}_1 & \mathbf{e}_2 & \mathbf{e}_3 \end{bmatrix}$ as shown in Fig.~\ref{fig:Perching Setup}. The quadrotor body frame, $\mathcal{B}$, is defined by $\begin{bmatrix} \mathbf{b}_1 & \mathbf{b}_2 & \mathbf{b}_3 \end{bmatrix}$. The target frame, $\mathcal{S}$, is represented by the landing pad axes $\begin{bmatrix} \mathbf{s}_1 & \mathbf{s}_2 & \mathbf{s}_3 \end{bmatrix}$ and identified in Fig.~\ref{fig:Perching Setup} with an Apriltag~\cite{olson2011tags}. Similarly, the position of the quadrotor's center of mass in the inertial frame is $\bm{x} = \begin{bmatrix} x & y & z \end{bmatrix}^{\top}$, and the target's center as $\bm{s} = \begin{bmatrix} s_x & s_y & s_z \end{bmatrix}^{\top}$. The perching problem requires the vehicles to plan and execute a feasible trajectory in a predefined time frame, $t \in [t_0,t_f]$, such that both $\mathcal{B}\equiv\mathcal{S}$ and $\bm{x} = \bm{s}$ are reached at $t=t_f$. This is achieved in a loop structure demonstrated in Fig.~\ref{fig:blockdiagram}. First, the vehicle visually locates the target and estimates the relative transformation from the $\mathcal{B}$ to $\mathcal{S}$ frames (i.e., relative position $\mathbf{p}_{\mathcal{S}}^{\mathcal{B}}\in\mathbb{R}^3$ and orientation $\mathbf{R}_{\mathcal{S}}^{\mathcal{B}}\in SO(3)$). Second, the relative configuration information is incorporated at the planning and control levels to generate and execute trajectories that are feasible.
The proposed setup and approach is shown in Fig.~\ref{fig:blockdiagram}. The quadrotor system dynamic model in the inertial frame $\mathcal{I}$ is 
\begin{equation}\label{eqn:model}
\begin{split}
&\dot{\bm{x}} = \bm{v}, \dot{\bm{v}} = \boldsymbol{a}, m\boldsymbol{a} = \mathbf{R}\tau \mathbf{e}_3 - mg\mathbf{e}_3, 
\\
&\Dot{\mathbf{R}} = \mathbf{R}\hat{\bm{\Omega}}, \mathbf{J}\Dot{\bm{\Omega}} + \bm{\Omega}\times \mathbf{J}\bm{\Omega} = \mathbf{M},
\end{split}
\end{equation}
where $\bm{x}, \bm{v}, \boldsymbol{a} \in {\mathbb{R}} ^3$ are the position, velocity, and acceleration of the quadrotor's center of mass in Cartesian coordinates with respect to the inertial frame $\mathcal{I}$, $\mathbf{R}$ represents the orientation of the quadrotor with respect to $\mathcal{I}$.  $\mathbf{\Omega} \in {\mathbb{R}} ^3$ is the angular velocity of the quadrotor with respect to $\mathcal{B}$, $m\in\mathbb{R}$ denotes the mass of the quadrotor, $\mathbf{J} \in {\mathbb{R}}^{3 \times 3}$ represents its inertial matrix with respect to $\mathcal{B}$, $g = 9.81$~m/s$^2$ is the standard gravitational acceleration, $\mathbf{M}\in\mathbb{R}^3$ is the total moment with respect to $\mathcal{B}$, $\tau\in {\mathbb{R}} $ represents the total thrust to the quadrotor, and the $\hat{\cdot}$ represents the mapping such that $\hat{\mathbf{a}}\mathbf{b} = \mathbf{a} \times \mathbf{b}, \forall \mathbf{a},\mathbf{b} \in \mathbb{R}^{3}$.

To achieve aggressive maneuvers, we apply a nonlinear geometric controller that was leveraged from our previous work~\cite{LoiannoRAL2017} to achieve agile flight in an indoor environments.  
where $\mathbf{K}_R, \mathbf{K}_{\Omega}, \mathbf{K}_x, \mathbf{K}_{v}\in\mathbb{R}^{3\times3}$  represent the feedback gains for the errors in orientation, angular velocity, position and velocity respectively as positive definite diagonal matrices. Thrust $\tau$ and moment $\mathbf{M}$ are the control inputs selected as

\begin{equation}\label{eqn:control}
\begin{split}
\tau &= \left(-\mathbf{K}_x \mathbf{e}_x - \mathbf{K}_{v} \mathbf{e}_{v} + mg\mathbf{e}_3 + m \ddot{\bm{x}}\right) \cdot \mathbf{R}\mathbf{e}_3=\mathbf{f}\cdot \mathbf{R}\mathbf{e}_3, \\
\mathbf{M} &= -\mathbf{K}_R \mathbf{e}_R - \mathbf{K}_{\Omega} \mathbf{e}_{\Omega} + \bm{\Omega} \times \mathbf{J}\bm{\Omega}  \\ &\hspace{50pt}-\mathbf{J}\left(\hat{\bm{\Omega}}\mathbf{R}^{\top}\mathbf{R}_C\bm{\Omega}_C - \mathbf{R}^{\top}\mathbf{R}_C\dot{\bm{\Omega}}_C\right), 
\end{split}
\end{equation}

$\mathbf{e}_R, \mathbf{e}_{\Omega}, \mathbf{e}_x, \mathbf{e}_v\in\mathbb{R}^3$ are the orientation, angular velocity, position and velocity error vectors this are detailed in works ~\cite{McLamrock, Loianno_IROS2015}, and the $*_C$ are the command or desired values obtained from the planning algorithm as shown in Section~\ref{sec:FoV_const}.
To plan a vehicle's motion in the $\mathcal{I}$ frame with respect to the target, the robot needs to localize itself with respect to the inertial frame $\mathcal{I}$ as well as with respect to the target.

Compared to our previous work~\cite{MAO_visaul_perch}, the robot continuously exploits the increased accuracy on target localization approaching the target to refine the estimate of the relative configurations from the $\mathcal{B}$ to $\mathcal{S}$ frames. Therefore, as shown in Fig.~\ref{fig:blockdiagram}, the system loops back with an updated target and quadrotor localization to replan trajectories in an active fashion
with an updated target and quadrotor until the quadrotor has intercepted the target as described in the following.

\section{Active Visual Perching}~\label{sec:AVP}
\begin{figure*}[!t]
    \centering
    \includegraphics[width=\linewidth]{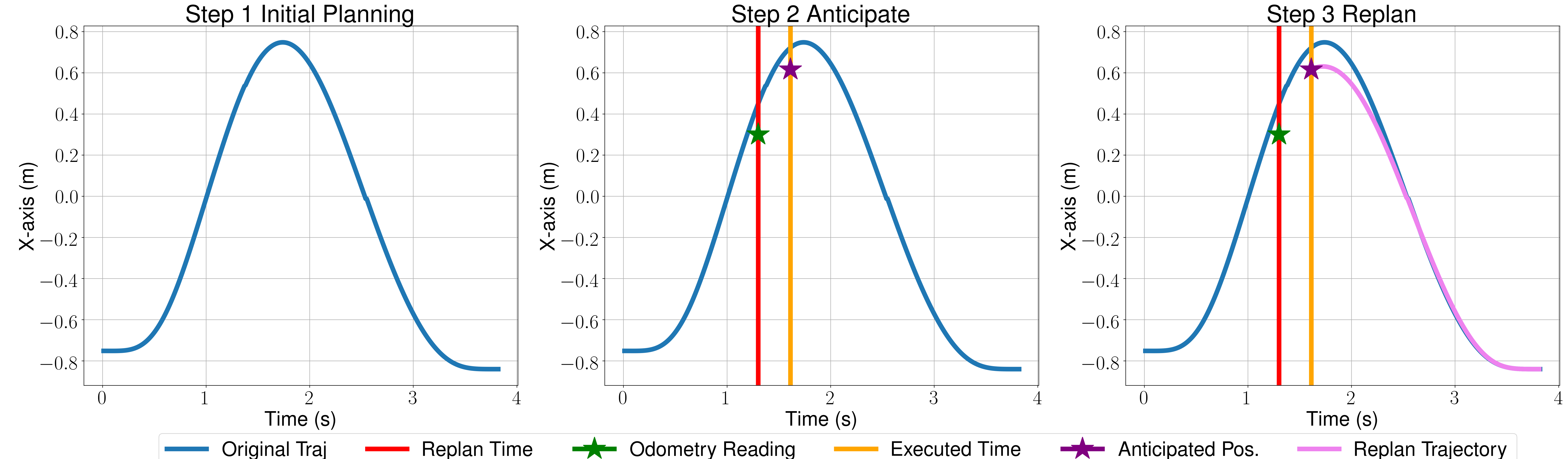}
    \caption{Replanning steps for one axis. This The horizontal axis is time, and the vertical is the X-axis distance. Blue represents the original trajectory. Red is when we start replanning. Orange is when we think the trajectory will take place. Green star is the current odometry. Purple star is the anticipated position. The pink trajectory is the new replanned trajectory.}
    \label{fig:replan}
\end{figure*}

The goal of Active Visual Perching (AVP) is to be able to plan new trajectories mid-flight to increase robustness to sensor and control disturbances and noises during aggressive flight by exploiting increased target localization accuracy and precision once the vehicle approaches the target for perching. Compared to our previous work~\cite{MAO_visaul_perch}, the active visual perching strategy imposes two additional steps in the aforementioned planner. First, since the vehicle is moving  while planning and the planning procedure takes a specific amount of time, we have to anticipate our flat output and corresponding time derivatives in the future when we swap the new trajectory. Second, we need to enforce the target visibility to guarantee and facilitate future replanning. This translates to a FoV constraint along sparsely discretized points of your trajectory. The FoV constraint formulation and implementation as well as the additional constraints that ensure the feasibility of the perching maneuver are detailed in Section~\ref{sec:FoV_const}. In  Fig.~\ref{fig:replan}, we visualize the replanning procedure for one Cartesian axis, $x$. The same procedure is valid for the other axes. Our procedure can be broken into three steps: initial planning, anticipation of the next position, and replanning from the anticipated time and flat output along with corresponding  derivatives.
First, we have our initial trajectory formulated in step 1 with a set amount of time for perching.
Similarly to~\cite{MAO_visaul_perch}, we model the trajectory on the set of vehicle's flat output space,  $\{\bm{x},\psi\}=\{x,y,z,\psi\}$ where $\psi$ is the yaw angle of the quadrotor or the rotation around the $\mathbf{b}_3$ axis. Each flat output trajectory is independent and represented using a polynomial spline. We give an example of a single dimension's spline defined as 
\begin{equation} \label{eqn:poly_spline}
\begin{split}
P(t) &= 
     \begin{cases}
       \text{$p_{1}\left(t-t_0\right)$} &\quad\text{if $t 	\in \left[t_0,t_1\right]$}\\
        \text{$p_{2}\left(t-t_{1}\right)$}&\quad\text{if $t 	\in \left[t_{2},t_{1}\right]$}\\
        \hspace{10pt}\vdots\\
        \text{$p_{f}\left(t-t_{f-1}\right)$} &\quad\text{if $t 	\in \left[t_{f-1},t_{f}\right]$}\\
     \end{cases},\\
 p_{i}(t) &= \sum_{n=0}^{N} c_{ni}t^{n}~,~i = 1, \cdots,f,
     \end{split}
\end{equation}
where $p_{i}$ represents the $i^{th}$ polynomial spline making up the full trajectory of $P$, $c_{ni}\in\mathbb{R}$ is the $n^{th}$ coefficient of $p_{i}$, and $N$ represents the polynomial order chosen for each spline. For our polynomial spline, we generate a reasonable guess at the required time needed for each spline. This formulation allows us to declare arbitrary constraints on the quadrotor's flat outputs and corresponding derivatives as linear constraints for any arbitrary time or time range as defined in~\cite{MAO_visaul_perch}. We minimize the squared norm of the $j^{th}$ derivative defined as 
\begin{equation} \label{eqn:cost_integral}\int_{t_{0}}^{t_{f}}\left\lVert\frac{d^{j} P(t)}{d t^{j}}\right\rVert^{2} dt.
\end{equation}

As in our previous work~\cite{MAO_visaul_perch}, we formalize this problem as a Quadratic Programming (QP) problem on the polynomial coefficients $\mathbf{c}$ defined as 
\begin{equation}\label{eqn:op_problem}
\begin{aligned}
\min_{\mathbf{c}} \quad & \mathbf{c}^T \mathbf{Q} \mathbf{c},\\
\textrm{s.t.} \quad & \mathbf{A} \mathbf{c}=\mathbf{b},\\
& \mathbf{y} \leq \mathbf{G} \mathbf{c} \leq \mathbf{z},   \\
\end{aligned}
\end{equation}
where the matrix ${\mathbf{Q}}$ is derived from the cost function in eq.~(\ref{eqn:cost_integral}), and the matrices $ \mathbf{A}, \mathbf{b}$ are derived from the equality constraint declared on flat outputs and corresponding derivatives for some arbitrary times, and  $ \mathbf{y},\mathbf{G}, \mathbf{z}$ on inequality constraints as defined in \cite{MAO_visaul_perch}.
Next in step 2, we detect our current position which is represented by the green star and $t_{rp}$, replan starting time represented by the red vertical line. Then, to take into account the overall replanning time duration on the vehicle's motion, we incorporate the quadrotor's flat output and corresponding derivatives with the spatio-temporal information by inferring the displacement of the quadrotor once the replanning ends as
\begin{equation}\label{eqn:anti}
\begin{split}
  \bm{x}_0 &= \bm{x}_n+\mathbf{P}(t_{ex})-\mathbf{P}(t_{rp}),\\
\frac{d\bm{x}_0}{dt} &= \frac{d\bm{x}_n}{dt}+\frac{d \mathbf{P}(t_{ex})}{dt}-\frac{d\mathbf{P}(t_{rp})}{dt},\\ 
&\vdots\\
\frac{d^j\bm{x}_0}{dt^j} &= \frac{d^j\bm{x}_n}{dt^j}+\frac{d^j \mathbf{P}(t_{ex})}{dt^j}-\frac{d^j\mathbf{P}(t_{rp})}{dt^j}.  \\ 
\end{split}
\end{equation}
It should be noted that we use $\mathbf{P}$ to refer to the trajectory on the full flat output states as opposed to $P$ for one flat output. The yaw has not been included for simplicity of notation.  The quadrotor's position is represented by the purple star whereas the orange line represents the anticipated time of execution, $t_{ex}$. The anticipation $\bm{x}_0$  is the spatial displacement that would occur in the originally planned trajectory and adding it to the odometry position $\bm{x}_n$. This is repeated for velocity, acceleration, and up to the fourth order derivative as shown in eq.~(\ref{eqn:anti}). In our final step, we plan a new perching trajectory with the anticipated flat output including all the relevant constraints detailed in Section~\ref{sec:replan constr}. The process is repeated until target interception.

\section{Active Visual Perching Constraints}\label{sec:replan constr}
In this section, we detail the specific constraints in eq.~(\ref{eqn:op_problem}) of the proposed AVP to intercept the target at the endpoint and maintain line of sight to the target with the robot front camera. We also derive the additional constraints required to intercept the target and show how the choice of the target's impact velocity influences the spatio-temporal target visibility during the maneuver. The above constraints represent inequality and equality constraints in the proposed optimization problem in  eq.~(\ref{eqn:op_problem}). Finally, we detail the procedure to guarantee that the planning constraints are met.  


\subsection{Field of View Constraint}\label{sec:FoV_const}
\begin{figure}[!t]
    \centering
    \includegraphics[width=\linewidth]{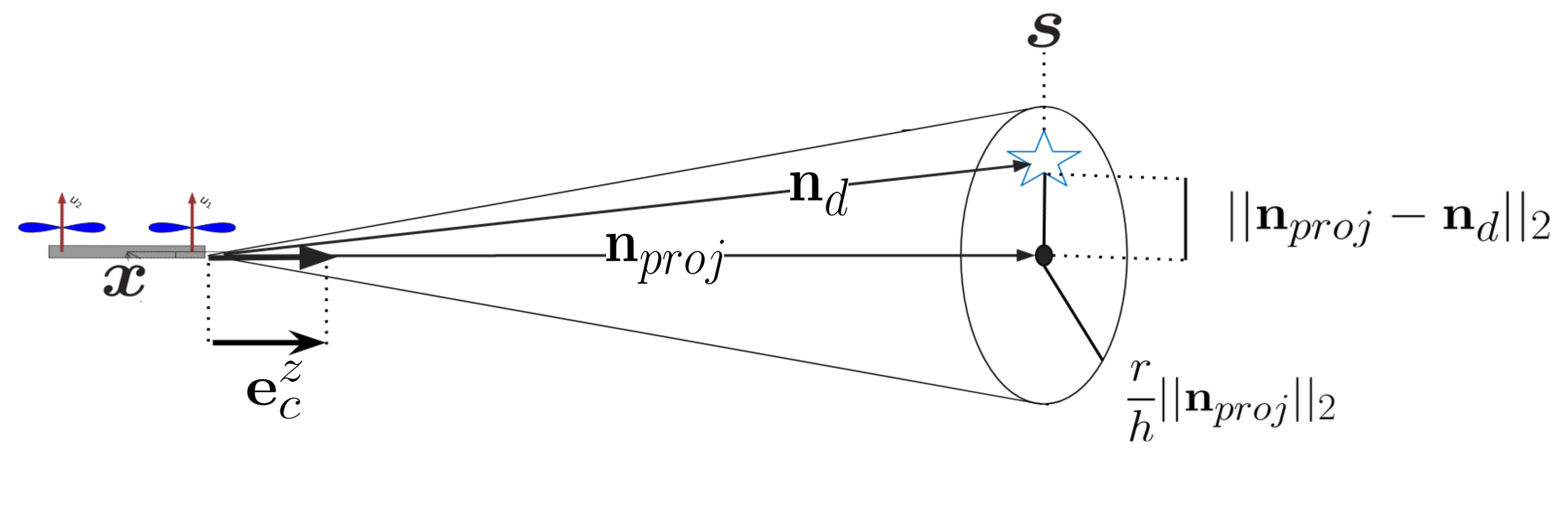}
    \caption{FoV constraint representation. $\bm{x}$ is the quadrotor's position. $\bm{s}$ and the star represents the target's position}
    \label{fig:FoV}
\end{figure}

On approach to the target, target localization improves as long as the target is in the FoV as quantified in Fig.~\ref{fig:target_localization}. This motivates a FoV constraint to be imposed onto the trajectory planning to maintain line of sight, so that the proposed AVP algorithm can exploit better localization. Our goal with this FoV constraint is to derive a linear constraint that is compatible with the optimization problem described in eq.~(\ref{eqn:op_problem}) and ensures that the next time the quadrotor needs the target localization for planning it is in the correct position and orientation to view the target. Inspired by~\cite{PCMPC}, we formulate our FoV constraint as a cone projected from the center of the drone's front camera. We refer to Fig.~\ref{fig:FoV} for a visualization. The relevant quantities for this constraint will be represented in the frame $\mathcal{I}$. 

First, $\bm{x}$ and $\bm{s}$ represent the drone and perching target's location. $\mathbf{n}_d = \bm{s}-\bm{x}$  represents the vector from the drone to the target. $\mathbf{e}^Z_c$ represents the unit vector projected from the camera's center. $\mathbf{n}_{proj}$ is the projection of $\mathbf{n}_{d}$ on to $\mathbf{e}^Z_c$. Finally, the ratio $\frac{r}{h}$ represents the ratio between the FoV cone's radius and height. To solve for $\mathbf{n}_{proj}$, we take the projection of $\mathbf{n}_d$ on to $\mathbf{e}^Z_c$, such that  the $\mathbf{e}^Z_c$ direction is maintained. This vector is $\mathbf{n}_{proj}= (\mathbf{n}_d^T \mathbf{e}^Z_c) \mathbf{e}^Z_c$. We can then define the FoV as a circle with a radius $\frac{r}{h}||\mathbf{n}_{proj}||_2 $ and center at $\mathbf{n}_{proj}$. Finally, we wish to enforce the target to be inside the circular FoV. This finalizes our constraint as
\begin{equation}\label{eqn:fov_nonlinear}
||\mathbf{n}_d-\mathbf{n}_{proj}||_2\leq\frac{r}{h}||\mathbf{n}_{proj}||_2.
\end{equation}

To apply our nonlinear FoV constraint, eq.~(\ref{eqn:fov_nonlinear}), to our optimization problem described in eq.~(\ref{eqn:op_problem}), we use a Taylor series expansion to linearize the inequality. Doing this linearization  represents an advantage from a computational perspective compared to employing nonlinear methods  as we show in Section~\ref{sec:results}.

\begin{figure*}[t]
    \centering
    \includegraphics[width=\linewidth]{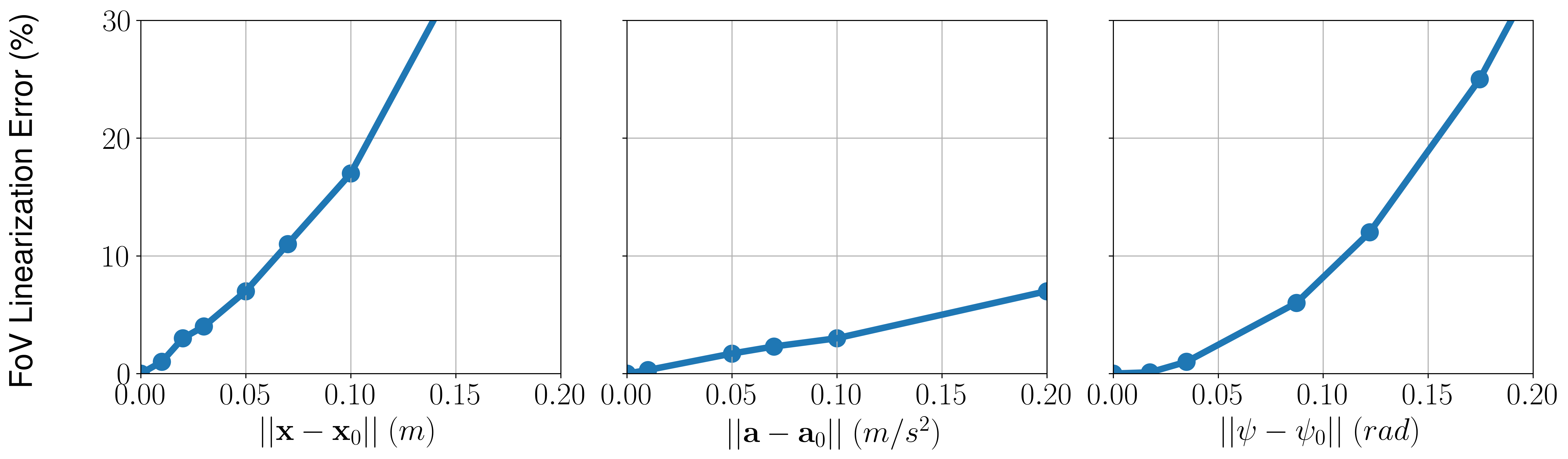}
    \caption{The linearization error eq.~(\ref{eq:FoV_linear}) of the field of view as a disturbance from the linearization zone. All plots share the same y-axis scale.}
    \label{fig:FoV_Linearization}
        \vspace{-10pt}
\end{figure*}

The constraint is a function of position, yaw, and acceleration. In other words, we can formulate eq.~(\ref{eqn:fov_nonlinear}) as $f(\bm{x},\psi,\mathbf{a})\leq g(\bm{x},\psi,\mathbf{a})$. The Taylor series expansion of this constraint linearized around $(\bm{x}_0,\psi_0,\mathbf{a}_0)$ becomes

\begin{equation} \label{eq:FoV_linear}
\begin{split}
&\left(\nabla_f\left(\bm{x}_0,\psi_0,\mathbf{a}_0\right)-\nabla_g\left(\bm{x}_0,\psi_0,\bm{a}_0\right)\right)\begin{bmatrix}
\bm{x}\\ 
\psi\\ 
\bm{a}
\end{bmatrix} \leq \\ &g(\bm{x}_0,\psi_0,\bm{a}_0)-f(\bm{x}_0,\psi_0,\bm{a}_0) +\\
&\left(\nabla_f\left(\bm{x}_0,\psi_0,\bm{a}_0\right)-\nabla_g\left(\bm{x}_0,\psi_0,\bm{a}_0\right)\right)\begin{bmatrix}
\bm{x}_0\\ 
\psi_0\\ 
\bm{a}_0
\end{bmatrix},
\end{split}
\end{equation}
where $\nabla_f$ and $\nabla_g$ represent the gradients of $f(x)$ and $g(x)$ with respect to $(\bm{x},\psi,\bm{a})$. 
For our constraint, we select our linearization point, $(\bm{x}_0,\psi_0,\bm{a}_0)$, as the next values from the next point we replan from to ensure that the target is still in view. These values $(\bm{x}_0,\psi_0,\bm{a}_0)$ are calculated using eq.~(\ref{eqn:anti}). Additionally, we impose an additional constraint on our position, yaw, and acceleration during replanning such that \begin{equation}\label{eq:FoV_region}
\begin{split}
||\bm{x} - \bm{x}_0||_2  &\leq 0.05,\\
||\bm{a} - \bm{a}_0||_2  &\leq 0.2,\\ 
||\psi - \psi_0||_2   &\leq 0.1,
\end{split}
\end{equation}
to ensure bounded linearization error as shown in Fig.~\ref{fig:FoV_Linearization} that details the Taylor series expansion error as a function of $||\bm{x} - \bm{x}_0||_2 $, $||\psi - \psi_0||_2 $, and $||\bm{a} - \bm{a}_0||_2 $. 
From this result, the Taylor series represents a good approximation of the original nonlinear constraint as long as the eq.~(\ref{eq:FoV_region}) is satisfied. We repeat the linear FoV constraint eq.~(\ref{eq:FoV_linear}) for $n_p$ points spaced equally along the remaining trajectory. Each constraint,  eq.~(\ref{eqn:fov_nonlinear}), is linearized around  the predicted value of the previously planned trajectory. This is similar to other nonlinear optimizers such as Sequential Quadratic Programming (SQP) where the previous iteration is used as linearization points for the next nonlinear optimization iteration.  

\subsection{Perching Physical and Perception Constraints}~\label{sec:perchung_physicalconst}
Inspired by our previous work ~\cite{loianno_est}, we exploit the nonholonomic properties of the quadrotor and relate $\Ddot{\bm{x}}$ and its orientation $\mathbf{R}$. From the translational dynamics eq.~(\ref{eqn:model}), the thrust vector is a function of the quadrotor's acceleration. 
\begin{equation}\label{eqn:thrust}
 \tau  = m||\Ddot{\bm{x}} +g\mathbf{e}_3||. 
\end{equation}
Because thrust can only be actuated on the plane normal to the propeller blades, the generated force is solely on $\mathbf{b}_3$ axis of the body frame of a quadrotor. From here we can derive that $\mathbf{b}_3$ should be
\begin{equation}\label{eqn:b3}
\mathbf{b}_3  = \frac{\Ddot{\bm{x}} +g\mathbf{e}_3}{||\Ddot{\bm{x}} +g\mathbf{e}_3|| }.
\end{equation}
To achieve successful perching, the quadrotor's body frame, $\mathcal{B}$, must align with the target's frame, $\mathbf{R}_{\mathcal{S}}$, meaning $\mathbf{b}_3=\mathbf{s}_3$ at the trajectory's end time, $t_f$. The $\mathbf{s}_3$ direction on the inclined surfaces is extracted from the last column of the matrix $\mathbf{R}_{\mathcal{S}}$. This orientation $\mathbf{R}_{\mathcal{S}}$ is combining the target localization with respect to the body frame and the on-board localization of the body frame with respect to the inertial frame. Using eq.~(\ref{eqn:b3}), we can declare this constraint as an acceleration as
\begin{equation} \label{eqn:acceleration constraint}
\Ddot{\bm{x}}(t_f) = \alpha\mathbf{s}_3 -g\mathbf{e}_3,
\end{equation}
where $\alpha=||\Ddot{\bm{x}}(t_f) +g\mathbf{e}_3||\in\mathbb{R}$ corresponds to the thrust's norm. The planned $\Ddot{\bm{x}}$ is then set as a constraint for our optimization defined in eq.~(\ref{eqn:op_problem}). We declare our quadrotor with the Z($\psi$)-Y($\theta$)-X($\phi$) convention to construct the rotation from flat outputs as derived below. 
To control the other two axis of the vehicle, $\mathbf{b}_1, \mathbf{b}_2$, we relate our desired $\mathbf{R}_C$ in eq.~(\ref{eqn:control}) to a desired yaw angle $\psi_{des}$ as
\begin{equation}
\begin{split}
\mathbf{R}_C=& \begin{bmatrix} 
\mathbf{b}_{1,C} & \mathbf{b}_{2,C} & \mathbf{b}_{3,C},
\end{bmatrix},\\
  \mathbf{b}_{1,C} =& \frac{\mathbf{b}_{2,des} \times \mathbf{b}_{3}}{\vert\vert{\mathbf{b}_{2,des} \times \mathbf{b}_{3}}\vert\vert},~~~
  \mathbf{b}_{2,C} = \mathbf{b}_{3} \times \mathbf{b}_{1},\\~\nonumber
    \mathbf{b}_{2,des} =& \begin{bmatrix} -\sin\psi_{des}, & \cos\psi_{des}, & 0 \end{bmatrix}^\top,~\nonumber
  \mathbf{b}_{3,C} = \frac{\mathbf{f}}{\vert\vert{\mathbf{f}}\vert\vert}.\nonumber
  \label{eq:R_des}
\end{split}
\end{equation}

For perching on a planar surface, this $\psi_{des}$ can be any angle that does not induce a singularity in eq.~(\ref{eq:R_des}). In our construction of $\mathbf{R}_C$, a singularity occurs when the rotation around the roll or $\mathbf{b}_1$ axis is at $\pm 90^\circ$. Generally, we select it such that $\mathbf{b}_{2,des}$ is parallel to $\mathbf{s}_2$, which can be know from $\mathbf{R}_{\mathcal{S}}$. The commanded angular rate $\hat{\bm{\Omega}}_{C}$ in eq.~(\ref{eqn:control}) is then
\begin{equation}
  \hat{\bm{\Omega}}_{C} = \mathbf{R}_{C}^\top\dot{\mathbf{R}}_{C}.
\end{equation}


For practical purposes, most of the rotation should completed before target interception to stop the front of the quadrotor from bouncing off the target and failing to adhere to the target or stopping  the quadrotor's vital components such as the propellers from colliding with the landing pad.  This constraint is expressed by enforcing an additional acceleration range by a given $q$ tolerance in proximity of the target. 
To apply the inequality constraint to our optimization, we discretize the equation as
\begin{equation} \label{eqn:acceleration discrete ineq constraint}
\begin{split}
&\left(\alpha\mathbf{s}_3 -g\mathbf{e}_3\right) \leq \Ddot{\bm{x}}(t) \leq \left(1+q\right)\left(\alpha\mathbf{s}_3 -g\mathbf{e}_3\right),
\\
&\forall  t \in{\{t_f-t_k+j*dt\}},  \hspace{10pt} j \in \mathbb{Z} \And 0 \leq  j < \frac{t_k}{dt},
\end{split}
\end{equation}
where $dt$ is the sampling time of our trajectory planner and $t_k$ is the time prior to the impact which is user defined. Physically this inequality means that before impact the rotation of the quadrotor is close to the final orientation.
 \begin{figure}[t]
    \centering
    \includegraphics[width=\linewidth]{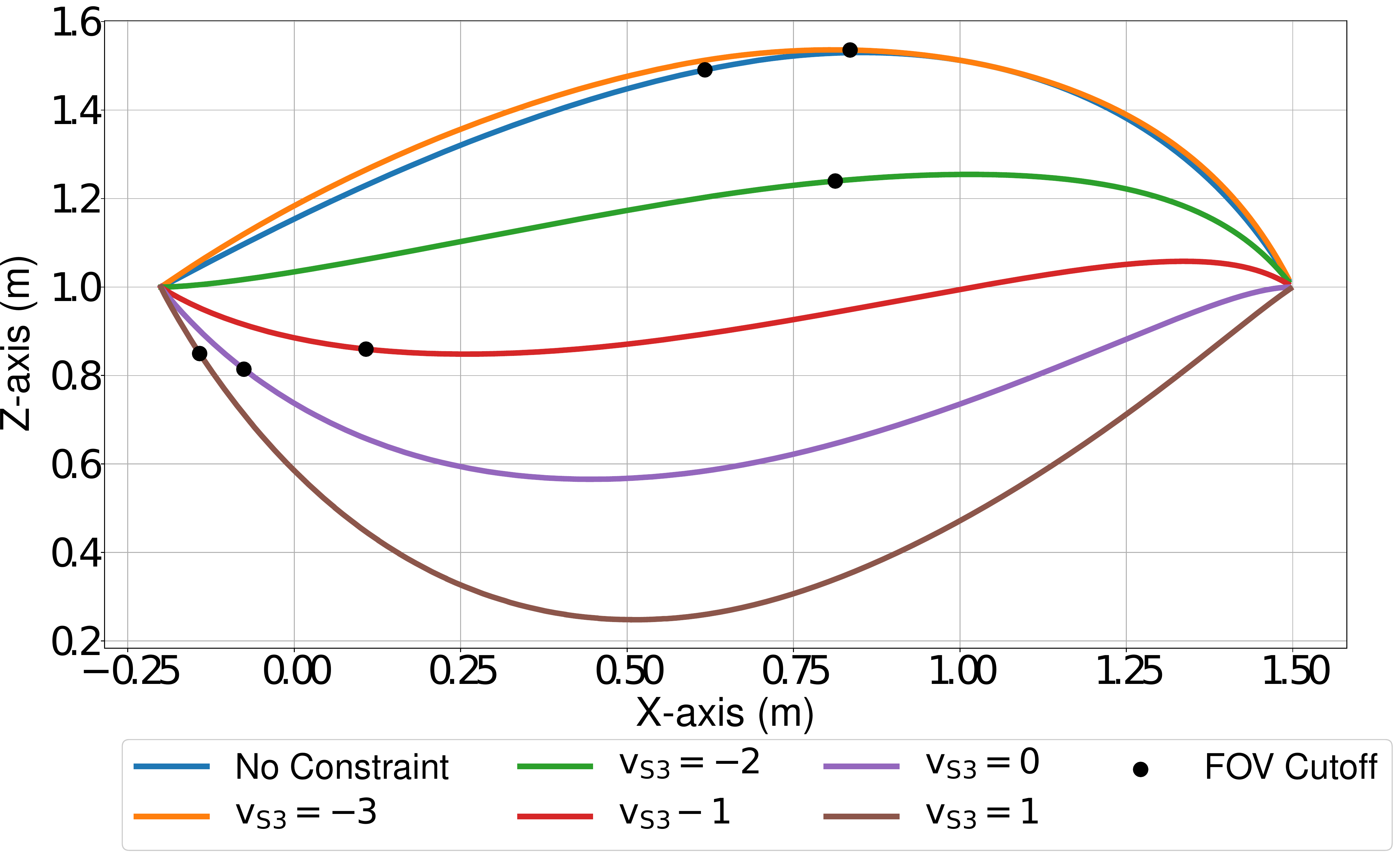}
    \caption{Visualization of each path for various $\mathbf{v}_{S3}$ during $90^\circ$ perching. $\mathbf{v}_{S1}=1 \si{m/s}$. The black dot is where the quadrotor loses sight of the target. Target's position is the endpoint of the trajectory. Velocities are in \si{m/s}.}
    \label{fig:FoV_path}
\end{figure}
\begin{table}[b]
\begin{center}
\caption{Percent of the time-Percent of distance ratio that target in its FoV as a function of $\mathbf{v}_{S1}$ and $\mathbf{v}_{S3}$\label{FoV_percent} during the perching maneuver. Velocities are in \si{m/s}.
}
\begin{tabular}{c c c c }
\hline
      &$\mathbf{v}_{S1} =0$ & $\mathbf{v}_{S1} =1$ &$\mathbf{v}_{S1} =2$\\
\hline
$\mathbf{v}_{S3} =\text{Null}$ & $65.5\% / 70.3\%$ & $78.1\%/ 63.9\%$ & $85.1\% / 65.1\%$\\
$\mathbf{v}_{S3} =-3$ & $68.9\%/ 72.5\%$ & $79.8\%/ 62.5\%$ & $85.6\%/ 65.8\%$ \\
$\mathbf{v}_{S3} =-2$ & $\mathbf{77.5\%} / \mathbf{83.7\%}$ & $85.1\%/75.1\%$ & $53.4\%/ 60.7\%$ \\
$\mathbf{v}_{S3} =-1$ & $29.8\% / 33.4\%$ & $41.3\%/ 48.3\%$ & $43.6\%/ 45\%$ \\
$\mathbf{v}_{S3} =0$ & $22.4\% / 10.4\%$ & $33.9\%/ 27.9\%$ & $38.5\%/ 34.2\%$ \\
$\mathbf{v}_{S3} =1$ & $17.8\% / 8.04\%$ & $27.0\%/ 20.4\%$ & $25.1\%/ 29.7\%$ \\
\hline
\end{tabular}
\end{center}
\end{table}
When perching at a very steep angle such as $90^\circ$ the endpoint velocity $\mathbf{v}$  can have a large effect on the perching path. This can either aid or greatly hinder the ability to see the target depending on if the quadrotor is ascending or descending to intercept. To characterize the effects of endpoint velocity on the FoV, we analyze how various endpoint velocities change the path and how well this trajectory maintains the target in the field of view for a trajectory. For our experiments, we consider two main factors: the percent of the time that the trajectory respects the FoV constraint and the ratio between the arc length where the FoV is respected over the full path's length. The qualitative and quantitative results of this study are demonstrated in Table~(\ref{FoV_percent}) and Fig.~\ref{fig:FoV_path}. Here, we consider mainly the following two components  $\mathbf{v}_{S1} = \mathbf{v}\cdot \mathbf{S}_1$ and $\mathbf{v}_{S3} = \mathbf{v}\cdot \mathbf{S}_3$, where $\mathbf{v}_{S1}$ is the velocity normal to the landing pad and $\mathbf{v}_{S3}$ is the velocity parallel to the landing pad. For the experiments in Table~\ref{FoV_percent}, we set the same time for each of the trajectories. From Fig.~\ref{fig:FoV_path}, we notice the effect of varying $\mathbf{v}_{S3} $. As $\mathbf{v}_{S3} $ approaches negative infinity, the maximum height is higher and for the reverse, as $\mathbf{v}_{S3} $ approaches infinity the minimum height of the path dramatically decreases. $\mathbf{v}_{S1}$ does not influence the maximum height or dip of perching but instead influences where the maximum or minimum dip happens. The faster the normal velocity is the closer to the start the hump will occur. In a generic use case, $\mathbf{v}_{S1}$ will also need to be tuned to meet the perching mechanism. However, for our mechanism, any non-zero $\mathbf{v}_{S1}$ is sufficient to achieve perching. We employ the combination of $\mathbf{v}_{S1} $ and $\mathbf{v}_{S3} $ based on our simulation results below to optimize the percent of the trajectory that respects the FoV constraint. Each of the above paths is generated assuming the FoV inequality condition is active for linearization of eq.~(\ref{eq:FoV_linear}) considering $n_p=8$ points in the future trajectory outlook.
\subsection{Perching Actuator Constraints}~\label{sec:act_constraints}
Once the optimization problem in eq.~(\ref{eqn:op_problem}) is solved, we must ensure that the actuator constraint is met
\begin{equation}\label{eqn:global constraint}
\begin{split}
\tau_{min}^2 \leq \left\lVert m\Ddot{\bm{x}}+mg \mathbf{e}_3 \right\rVert^2_2 \leq\tau_{max}^2
\end{split}
 \end{equation}
 where $\tau_{\text{min}}$ and $\tau_{\text{max}}$ are the minimum and maximum thrust respectively. This condition unlike the other previous conditions must be true for the entire trajectory flight time rather than a specific time frame. As a result, it is very difficult to incorporate the QP optimization problem described in eq.~(\ref{eqn:op_problem}). Direct incorporation in the optimization problem would increase the solving time. Using our previous work \cite{MAO_visaul_perch} which built on Sturm's theorem from \cite{sturm2009}, we perform efficient bounds check post-solving using the Global Bound Checking summarized in Algorithm~\ref{agl:bound}. Details about proof of this algorithm are provided in Section~\ref{sec:appendix}. This algorithm returns true if given some polynomial $H(t)$ and some constant $b$, then $H(t) < b$ for all $t \in [t_0,t_f]$. Otherwise, it returns false. Since we use polynomial splines to characterize our trajectory, the thrust is a polynomial. We apply the global bound checker after solving the QP optimization to check if eq.~(\ref{eqn:global constraint}) is respected. Should this constraint not be respected, we iteratively add time to each segment of the polynomial till the constraint is respected. This procedure takes place primarily after solving the initial planning. This is because the combined result of all the newly planned trajectories tends to follow a shorter trajectory and move at a lower average speed than a one-shot approach as we observe in the experimental section with Fig.~\ref{fig:Path_comp}. 
 \begin{algorithm}[!t]
\caption{{\sc Global Bound Checking (GBC)} \cite{MAO_visaul_perch} }
\label{agl:bound}
Returns true if $H\left(t\right) < b$ $\forall$ $t\in\left[t_0,t_f\right]$. $H(t)$ is any polynomial. $b \in \mathbb{R}$ is an upper bound
\begin{algorithmic}[1]
    \STATE Let $F(t) = H(t)- b$;
    \IF{$F\left(t_0\right) > 0 \hspace{10pt} \OR\hspace{10pt} F\left(t_f\right) > 0 $} 
        \RETURN FALSE
    \ENDIF
    \IF{STURM$\left(F(t), t_0,t_f\right) > 0 $} 
        \RETURN FALSE
    \ENDIF
    \RETURN TRUE
 \end{algorithmic}
\end{algorithm}

Additional constraints can be considered such as thrust rate, angular velocity, angular acceleration, moments, and moments rates to Algorithm 1 of our work. A derivation of these quantities can be found in Appendix~(\ref{sec:poly_bounds}). Therefore Sturm's theorem from \cite{sturm2009} can be applied to perform efficient bounds check. If we find the trajectory is infeasible, then we can add time to decrease the maximum of the above constraints similar to thrust and find a feasible trajectory. 

\subsection{Feasibility Conditions}
Finally, there exists the possibility that the visibility constraint will not be compatible with the perching constraints. In order to verify the existence of this case, we employ the Karush–Kuhn–Tucker (KKT) conditions. The KKT conditions for our quadratic programming defined by eq.~(\ref{eqn:op_problem}) with constraints specified in eqs.~(\ref{eq:FoV_linear}),~(\ref{eqn:acceleration constraint}), and~(\ref{eqn:acceleration discrete ineq constraint}) determine the solution if one exists. To formulate our KKT condition, we only consider inequality constraint of the form $\mathbf{y} \leq \mathbf{G} \mathbf{c}$. The general inequality constraint $\mathbf{G} \mathbf{c} \leq \mathbf{z}$ is equivalent to the linear constraint $-\mathbf{z} \leq -\mathbf{G} \mathbf{c}$, so the above formulation does not restrict the set of solvable problems.  Solving the KKT conditions involves creating two sets of Lagrange multipliers $\bm{\lambda}_1  \in \mathbb{R}^J$, $\bm{\lambda}_2 \in\mathbb{R}^K$, and a group of slack variables $\mathbf{s}\in\mathbb{R}^K$ where J and K are the number of equality and inequality constraints respectively. Based on these new variables, we can formulate the KKT conditions as  

\begin{equation}\label{eqn:KKT_cond}
\begin{aligned}
&\mathbf{Q} \mathbf{c}-\mathbf{A} \bm{\lambda}_1-\mathbf{G} \bm{\lambda}_2=0,
&\bm{\lambda}_1^\top\left(\mathbf{y}-\mathbf{G}\mathbf{c}\right)=0,\\
&\mathbf{A}\mathbf{c}-\mathbf{b}=0,
&\mathbf{G} \mathbf{c}-\mathbf{y}-\mathbf{s}=0,\\
&\bm{\lambda}_1,\bm{\lambda}_2,\mathbf{s}\geq0.\\
\end{aligned}
\end{equation}
We solve these conditions using the Primal-Dual Interior-Point method. Should an incompatibility between the FoV constraints and other objectives or constraints occur, we relax the problem by removing the FoV constraint and plan the trajectory by only incorporating the perching constraint.

\section{Results}~\label{sec:results}
In this section, we demonstrate the in simulation that our quadrotor's visibility constraints provide real improvements to the visibility of the target. Next, we show the improvement in target interception gained from using multiple detection on approach as opposed to one-shot planning Finally, we demonstrate that we can successfully perch on any target up to $90^\circ$ with only on-board sensing and computation.

\subsection{Setup}
\begin{figure}[t]
    \centering
    \includegraphics[width=\linewidth]{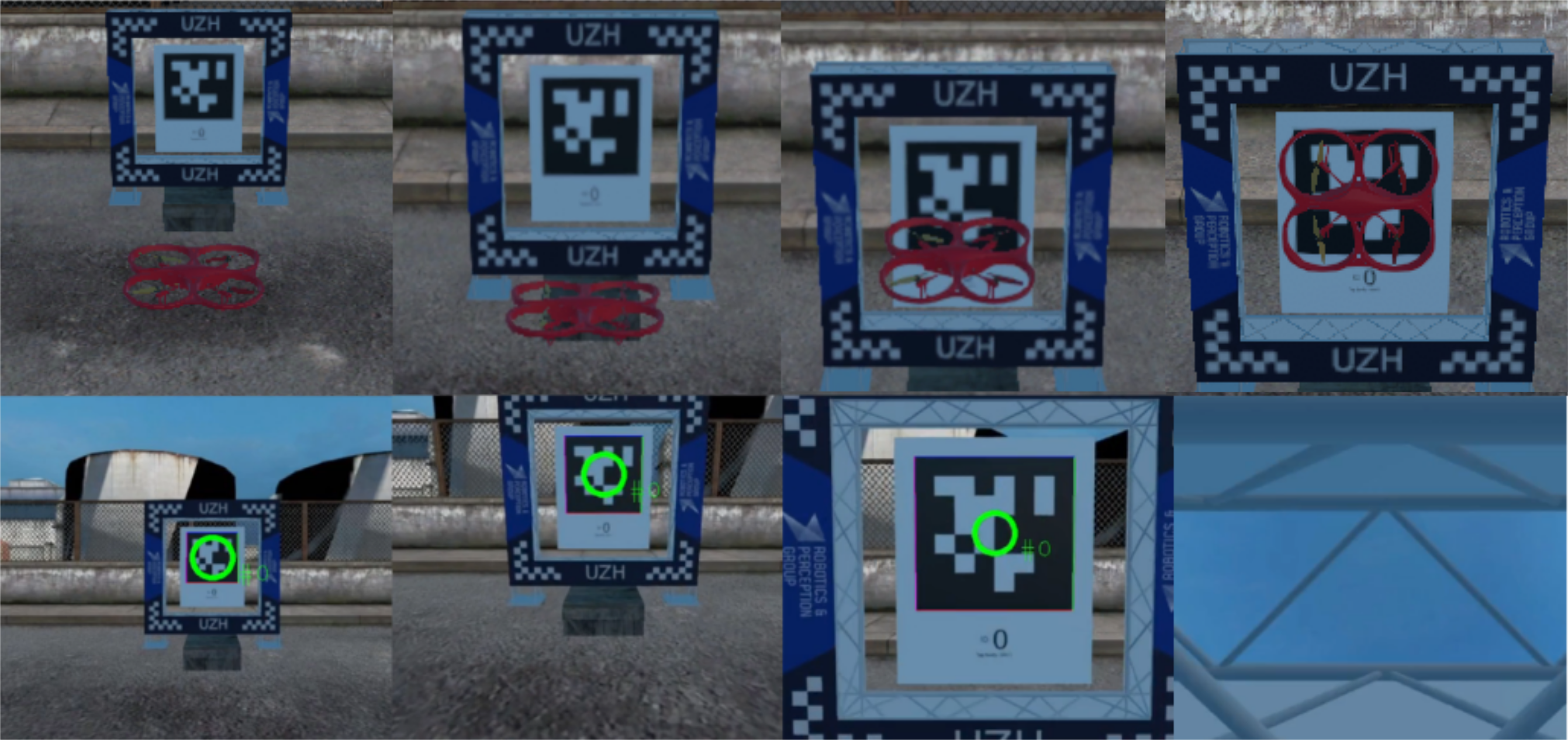}
    \caption{Virtual perching in simulation. The top $4$ images are the view of the quadrotor in the simulated environment. The bottom $4$ images display the view from the camera with tag detection. The green circle denotes the target detection. }
    \label{fig:flightmare_simulator}
        \vspace{-10pt}
\end{figure}

We deploy our approach both in simulation and real-world settings. For simulation, we consider a photo-realistic simulator based on a variation of Flightmare~\cite{song2020flightmare}. This simulator is capable of rendering both the quadrotor, environment, and front camera view as demonstrated in Fig.~\ref{fig:flightmare_simulator}. In our setup, Flightmare Unity only handles graphic rendering of both on-board cameras and the environment. This allows us to fully test and validate our framework without any setup variation in simulation. The simulated on-board camera images are fed back into our planning software to execute our AVP perching maneuver. The  physics simulation is built on our custom quadrotor simulator which takes in the output of our position controller and simulates the motion of the quadrotor (a simple flag sends the commands directly to a real platform).  The landing pad area is localized both in simulation and real world experiments using a single large Apriltag~\cite{olson2011tags}. Our landing pad is identified using the quadrotor's front camera that detects a single large Apriltag with a known offset from the main pad. In the real world, the pad is mounted on an adjustable desk that  controls the height. The adhesive is attached to an adjustable stand which allows variation in the surface angle. A sample perching sequence is demonstrated in simulation in Fig.~\ref{fig:flightmare_simulator}.
The real world experiments are conducted at the Agile Robotics and Perception Lab (ARPL) lab at New York University in a flying arena of dimensions  $10\times6\times4~\si{m^3}$. As shown in Fig.~\ref{fig:Perching Setup}, the platform
is a quadrotor running with a Qualcomm\textsuperscript{\textregistered} Snapdragon\textsuperscript{TM} board and 4 brushless motors. The Qualcomm\textsuperscript{\textregistered} Snapdragon\textsuperscript{TM} consists of a  Qualcomm\textsuperscript{\textregistered} Hexagon\textsuperscript{TM} DSP, Wi-Fi, Bluetooth, GPS, quad-core processor, an IMU, and two cameras: a downward facing
$160^\circ$ FoV and a front-facing $70^\circ$ FOV. For perching, we employ VELCRO\textsuperscript{\textregistered} material due to its simplicity and low cost in the ventral part of the vehicle. 
The software framework is developed in ROS\footnote{\url{www.ros.org}} Indigo on a Linux kernel. As specified in Section~\ref{sec:Methods}, localization with respect to the inertial frame $\mathcal{I}$ is performed based on the on-board Visual Inertial Odometry (VIO) system using the downward facing camera~\cite{loianno_est} upsampled with the IMU to $300~\si{Hz}$. The front-facing camera is used to localize the target at a rate of $10~ \si{Hz}$. A Vicon\footnote{\url{www.vicon.com}} motion capture system is used to record the ground truth data for comparison at $100~\si{Hz}$.  

In both simulation and real-world settings, we employ the same parameters and settings for our perching maneuver. We perform Active Visual Perching at a rate of $30$~\si{Hz} for replanning. In our setup, perching constraints are set as $\mathbf{v}_{S3} = -2~\si{m/s}$ and $\mathbf{v}_{S1} = 0.3~\si{m/s}$. The slight increase of $\mathbf{v}_{S1} = 0.3~\si{m/s}$ from the optimal $\mathbf{v}_{S1} = 0~\si{m/s}$ is because the VELCRO\textsuperscript{\textregistered} requires some forward momentum to attach itself.  We empirically identified that $0.3~\si{m/s}$ is a good forward momentum to attach to the landing pad. We selected a tolerance, $q=0.1$, sampling time, $dt=0.01~\si{s}$, time before impact, $t_k = 0.15~\si{s}$ and $\alpha = 4.0~\si{m/s^2}$ as the hyperparameters of eqs.~(\ref{eqn:acceleration constraint}) and (\ref{eqn:acceleration discrete ineq constraint}). We choose to minimize the $j=4$, snap norm, for the cost function in eq.~(\ref{eqn:cost_integral}). Regarding the FoV constraint in eq.~(\ref{eq:FoV_region}), we chose $8$ points to preserve a good trade-off between computation and ensuring the FoV constraint is enforced for most of the trajectory.


We  solve the optimization problem in eq.~(\ref{eqn:op_problem}) using the Object Oriented Quadratic Programming (OOQP) library \cite{OOQP}. Our optimizer concurrently solves the KKT conditions described in eq.~(\ref{eqn:KKT_cond}) and raises a flag simultaneously after solving if a solution exists or not. In case of no solution being found due to the incompatibility of the FoV constraint with other objectives or constraints, we drop the visibility constraints and resolve the problem considering only the perching constraints to guarantee the physical feasibility of the maneuver. 
The planning process takes $29$ \si{ms} on-board  our platform equipped with an Arm processing unit and $28$ \si{ms} on a novel NVIDIA Xavier NX. We also evaluate the computational time when leveraging recent nonlinear solvers like ACADOS~\cite{acados} with SQP~\cite{SQP}. Similar to \cite{ji2022real}, the proposed formulation using ACADOS solves an optimization problem considering a series of setpoints with a time difference of $1$ ms between each other as opposed to finding the polynomial coefficients as in our original approach. This formulation as a discrete time nonlinear optimization represents the way ACADOS treats these problems and is consistent with other works~\cite{ji2022real} in the field which have proposed a similar approach. While ACADOS can be used to solve for polynomial coefficents, it is unlikely to be an interesting comparison as ACADOS uses an optimizer called sequential quadratic programming  (SQP) which requires an inner QP optimizer where one of the given options is OOQP. ACADOS performs multiple iterations until the cost converges to below $5\mathrm{e}{-7}$ threshold.  The process takes $40$ \si{ms} on an NVIDIA\textsuperscript{\textregistered} Xavier NX. It is not possible to test ACADOS on-board our platform  because of multiple library/hardware incompatibilities  related to the specific arm architecture available on our quadrotor. We also execute the tests on a laptop equipped with an Intel i5-9300H CPU. We see both solvers nonlinear and linear give similar results $11$ \si{ms} and $13$ \si{ms} for OOQP and ACADOS respectively. The results are summarized in Table~\ref{tab:solving_time}. The computational time between linear quadratic programming and nonlinear when solving the problem off-board is quite similar, whereas on-board computationally constrained platforms, the time difference is larger. This is due to multiple iterations that the SQP performs which take a larger amount of time on-board computationally constrained platforms.

\begin{table}[!t]
\begin{center}
\caption {Computation time between different solvers running on various architectures.}
\begin{tabular}{ c c c c c c c}
    \hline
 Solver & Snapdragon~(\si{ms})&Xavier NX~(\si{ms})&Intel i5 ~(\si{ms})\\
\hline
          OOQP & $29$ & $28$ & $11$\\
          ACADOS & N/A & $40$ & $13$\\
 \hline
\end{tabular}
 \label{tab:solving_time}
\vspace{-10pt}
\end{center}
\end{table}


\subsection{Target Localization}
 \begin{figure}[!b]
    \centering
    \includegraphics[width=\linewidth]{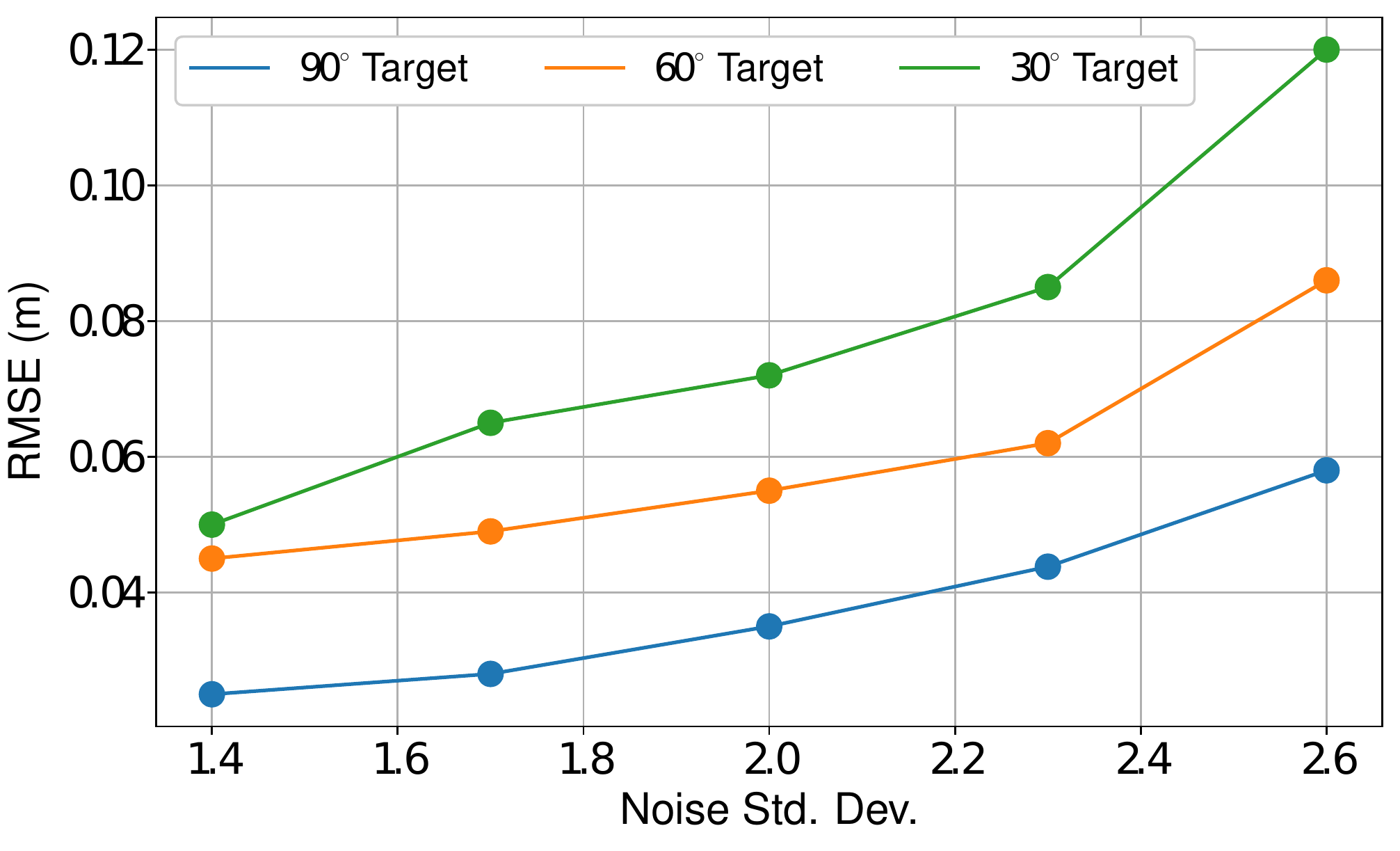}
    \caption{AprilTag localization error as a function of distance from the target at various inclines.}
    \label{fig:target_localization}
        \vspace{-10pt}
\end{figure}

As the distance between camera and target reduces, the vision based Apriltag target localization improves. In this experiment, we demonstrate this by flying a quadrotor level to the target and backing away horizontally to measure the localization error from the Apriltag and compare it to the ground truth pose values obtained using our Vicon motion capture system. This procedure is repeated at multiple inclines where the angle refers to the relative angle between the floor and the target. The results are recorded in Fig.~\ref{fig:target_localization}. As seen in Fig.~\ref{fig:target_localization} increasing the distance causes the target localization error to exponentially increase. This motivates the need for replanning in our system. Leveraging reduced localization error during the perching maneuver will improve perching consistency. The decrease of accuracy in target localization as the quadrotor moves further from the target is also successfully modeled in our simulator where we notice a similar accuracy drop off. 

\subsection{Active Visual Perching Evaluation}

\subsubsection{Simulation}

In our simulations, we perform a sequence of  studies to quantify the impact of various constraints and parameters. First, we validate that the constraints selected for planning a perching maneuver improve the target visibility in terms of both space and time. Second, we evaluate the accuracy of the anticipation step in the AVP. Finally, we demonstrate replanning improves the interception with the target by comparing one shot with AVP in simulation. The advantage of performing these experiments in simulation is that more variables are controlled and only the removed or altered component in the study causes performance differences between experiments.

\begin{table}[b]
\begin{center}
\caption{Ablation study of the role of the FoV and boundary conditions on the target visibility. The FoV constraint is represented by eq.~(\ref{eq:FoV_linear}), whereas $\mathbf{v}_{s1s3}$ refers to if both $\mathbf{v}_{s1}$ and $\mathbf{v}_{s3}$ are set or left floating. Time and space refer to the corresponding temporal or spatial FoV percentage during the perching maneuver. The distances such $1.5~\si{m}$, $3.5~\si{m}$, and $7.5~\si{m}$ refers to the distance the quadrotor to the landing pad along the $\mathbf{s}_{1}$ axis. }
\begin{tabular}{ c c c c c c c c c}

\hline
        \multicolumn{2}{c}{Constraint}     &  & \multicolumn{2}{c}{ 1.5~\si{m}~(\%)} & \multicolumn{2}{c}{ 3.5~\si{m}~(\%)} & \multicolumn{2}{c}{ 7.5~\si{m}~(\%)}\\
   \hline
      FoV &$\mathbf{v}_{s1s3}$ &  & Time & Space & Time & Space & Time & Space\\
 \cline{1-2} \cline{4-9} \ 
  \xmark & \xmark  &  & $65$&$44$& $71$&$52$& $80$&$74$\\
  \cmark & \xmark  &  & $\mathbf{72}$&$45$ & $82$&$53$& $\mathbf{89}$&$74$\\
  \xmark  & \cmark &  & $68$&$47$ & $80$&$76$ & $86$&$83$\\
 \cmark & \cmark &   &$70$&$\mathbf{53}$ & $\mathbf{82}$&$\mathbf{76}$& $87$&$\mathbf{88}$\\
 
\hline
\end{tabular}
\label{tbl:FoV_abalation}
\end{center}
\end{table} 
First, we must verify all relevant constraints: eq.~(\ref{eq:FoV_linear}),  $\mathbf{v}_{s1}$, and $\mathbf{v}_{s3}$ contribute to improving the visibility of the target. We first disable the constraints expressed by eq.~(\ref{eq:FoV_linear}) as well as $\mathbf{v}_{s1}$ and $\mathbf{v}_{s3}$ in the optimization. Our system then executes a perching trajectory and records how long in terms of both space and time the target is recognized with our tag detector. Spatial visibility refers to the percentage of the path traveled that the target is visible. Temporal visibility refers to the percentage of time in performing the maneuver the target is visible. We repeated this experiment with every combination of the constraints  eq.~(\ref{eq:FoV_linear}),  $\mathbf{v}_{s1}$ and $\mathbf{v}_{s3}$ at a distance of $1.5~\si{m}$, $3.5~\si{m}$, and  $7.5~\si{m}$ along the  $\mathbf{s}_{1}$ axis away from the target. In this case,  the vehicle was aligned with the target on the y-axis. The results are presented in Table~\ref{tbl:FoV_abalation}. The results show that as expected the FoV constraint eq.~(\ref{eq:FoV_linear}) is useful to enforce target spatio-temporal visibility. By carefully selecting $\mathbf{v}_{s1}$ and $\mathbf{v}_{s3}$,  we obtain a better spatial visibility of the target compared to the case when no constraints are active. Combining these two constraints, we can achieve a stronger spatio-temporal target visibility when perching without substantially comprising on either. 
Additional experiments are performed to verify that the visibility both spatially and temporally remains consistent for different starting positions. We changed the relative position of the quadrotor and executed a perching maneuver with all visibility constraints active. The spatio-temporal visibility of the target is reported in Table~\ref{tbl:FoV_lateral motion}. We notice that on average shifting the height of the quadrotor or moving it along the lateral does not have a major impact on the visibility constraint. In addition, in the aforementioned situations, the quadrotor is always able to successfully perch despite the positional shifts. We did not shift the height of the quadrotor for the $1.5~\si{m}$ case because the target starts outside the quadrotor's field of view in this case meaning we have no initial estimate to localize to or track before perching. This shows that our visibility constraints are fairly robust to different starting conditions.


\begin{table}[t]
\begin{center}
\caption{Temporal-spatial visibility of simulated perching. $1.5/3.5/7.5$ ~\si{m} refer to position of the quadrotor along the $\mathbf{s}_{1}$ axis. The - sign represents the case when the initial position does not allow the identification of the target.
}
\resizebox{0.95\columnwidth}{!}{
\begin{tabular}{c c c c c c c c c}

\hline
        \multicolumn{2}{c}{Axis} &  & \multicolumn{2}{c}{ ~1.5~\si{m}~(\%)} & \multicolumn{2}{c}{~3.5~\si{m}~(\%)} & \multicolumn{2}{c}{~7.5~\si{m}~(\%)}\\
   \hline
      $\mathbf{s}_{2}$~(\si{m}) &$\mathbf{s}_{3}$~(\si{m}) &  & Time & Space & Time & Space & Time & Space\\
 \cline{1-2} \cline{4-9} \ 
$0$ & $0$  &  & $70$&$53$& $82$&$71$& $87$&$88$\\
 $0.5$ & $0$ &   &$70$&$54$ & $81$ &$76$ & $85$&$83$\\

 $2$ & $0$ &   &$68$&$50$ & $82$&$75$& $86$&$85$\\
  $4$ & $0$ &   &-&- & $85$&$75$& $87$&$84$\\

  $-0.5$ & $0$ &   &$70$&$54$ & $79$&$76$& $84$&$83$\\
   $-2$ & $0$ &   &$69$&$50$ & $82$&$75$& $87$&$84$\\
  $-4$ & $0$ &   &-&- & $85$&$75$& $88$&$84$\\

  $0$ & $-0.8$  &  & -&- & $81$&$73$& $86$&$83$\\
  $0$ & $1$ &  & -&- & $80$&$73$ & $87$&$85$\\

\hline
\end{tabular}}
\label{tbl:FoV_lateral motion}
\end{center}
\end{table}
We also evaluate the prediction accuracy of  the anticipation, eq.~(\ref{eqn:anti}). In this experiment, we consider a perching maneuver in simulation at various distances and calculated the Root Mean Square Error (RMSE) between the first predicted point and the quadrotor position when switching to a new trajectory. The results are shown in Table~\ref{tab:anti_error}. There is not much error on average between the true pose and replanning on the order of \si{cm} showing that our prediction is quite accurate. We did not record the y-error in the following table since the quadrotor was aligned with the target and the error was negligible for anticipation or without anticipation. This experiment has also been conducted in the real world and a similar trend to simulation is observed. 
\begin{table}[!b]
\begin{center}

\caption {Study of anticipation error eq.~(\ref{eqn:anti}) in simulation. The distances $1.5/3.5/7.5~\si{m}$ refer to the how far the quadrotor is from the target on the $\mathbf{s}_1$ component. \label{tab:anti_error}}
\begin{tabular}{ c c c c c c c}
    \hline
  &\multicolumn{2}{c}{1.5 m}&\multicolumn{2}{c}{3.5 m}&\multicolumn{2}{c}{7.5 m}\\
\cline{2-7}\
            Axis & $x$ & $z$   & $x$ &  $z$   & $x$ & $z$ \\\hline
 Simul. RMSE~(\si{m}) & $0.014$   & $0.003$ & $0.02$   &$0.003$&$0.03 $  & $0.006  $\\
 Real RMSE~(\si{m}) & $0.015$   & $0.002$ & $0.03$   &$0.01$&$0.04 $  & $0.02  $\\

    \hline
\end{tabular}
\end{center}
\end{table}

\begin{figure}[t]
    \centering
    \includegraphics[width=1\columnwidth]{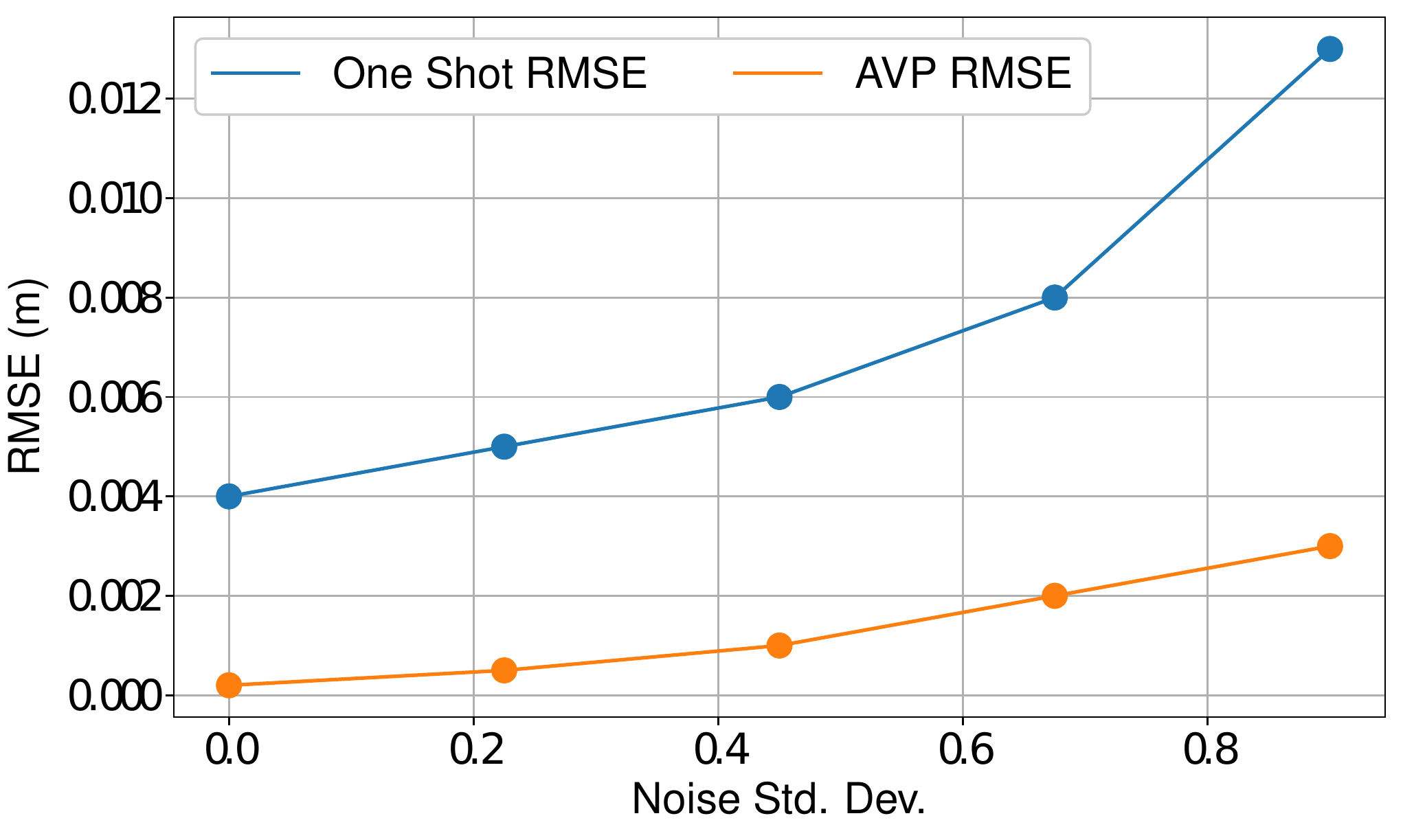}
    \caption{The control error as a function of the error standard deviation in simulation. The quadrotor was tasked with perching on a target $1.7$~m away at an incline of $90^\circ$.}
    \label{fig:Sensor_noise}
        \vspace{-10pt}
\end{figure}

Finally, we aim to study how AVP improves the perching maneuver with respect to one shot. In the first experiment, we artificially inject zero mean Gaussian noise with different variances ranging from $0-1$ in eq.~(\ref{eqn:control}) on the state estimation.  After injecting the noise in the state estimation, we set the quadrotor to perform a standard perching trajectory on a $90^\circ$ incline in simulation from a distance of $1.7$~\si{m}. This experiment is repeated for both a one-shot and AVP approach. The tracking RMSE between the planned and true path is reported in Fig.~\ref{fig:Sensor_noise}. The proposed AVP approach shows increased robustness to injected noise in terms of tracking error compared to the one shot approach. These trials were repeated $5$ times each. Additionally, we demonstrated that AVP can improve target interception by calculating the final perched RMSE between the center of the target and the quadrotor or $||\bf{x}-\mathbf{s}||$. This process is repeated for various distances as shown in Table~\ref{tab:sim_center_error}. We use an average of $5$ trials in simulation to calculate the average RMSE for the center error.  For shorter distances, the difference in error is not significant. However, if we increase the distance it becomes clear that the AVP can better localize the target with repeated plannings and intercept closer to the center especially at $7.5$~\si{m}. 
\begin{table}[!t]
\begin{center}
\caption {The error of the quadrotor's final position with respect to the target in terms of the two axes. $1.5/3.5/7.5~\si{m}$ refers to the initial starting distance of the quadrotor with respect to the target.}
\begin{tabular}{ c c c c c c c}
    \hline
 Axis&\multicolumn{2}{c}{$1.5~\si{m}$}&\multicolumn{2}{c}{$3.5~\si{m}$}&\multicolumn{2}{c}{$7.5~\si{m}$}\\
\cline{2-7}\
            & One Shot   & AVP & One Shot   & AVP& One Shot   & AVP\\\hline
            $\mathbf{s}_2~(\si{m})$ & $0.03$   & $0.02$ & $0.04$   & $0.02$ 
            & $0.1$   & $0.03$  \\
            $\mathbf{s}_3~(\si{m})$& $0.04$  & $0.05$ & $0.06$   & $0.05$  
            & $0.25$   & $0.09$  \\\hline
\end{tabular}
 \label{tab:sim_center_error}
\vspace{-10pt}
\end{center}
\end{table}

\subsubsection{Real World Experiments}
\begin{figure}[!ht]
    \centering
    \includegraphics[width=\linewidth]{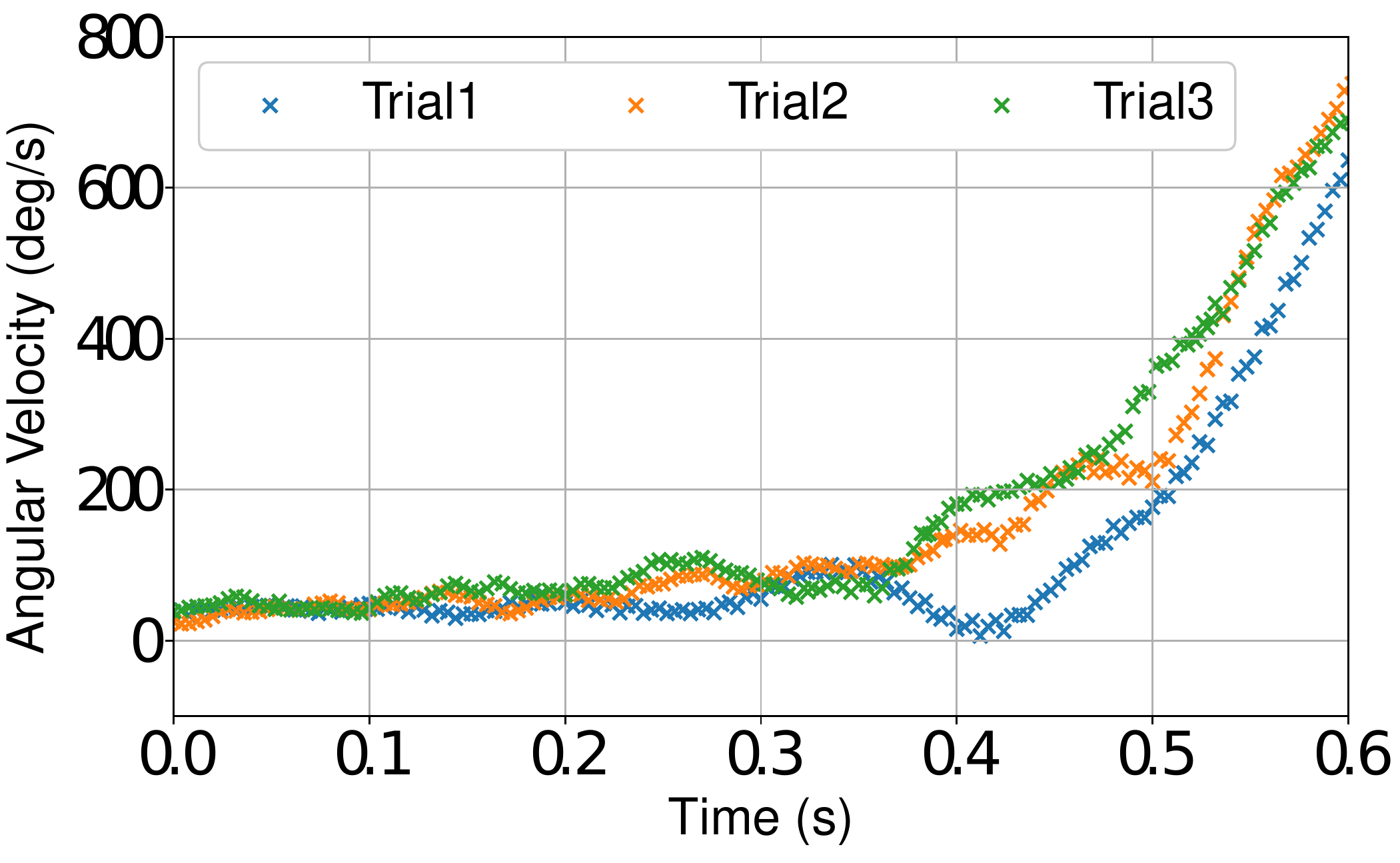}
    \caption{Angular rate along the $\mathbf{b}_2$ axis during a series of $90^\circ$ perching maneuvers. }
    \label{fig:angular_rate}
\end{figure}

We demonstrate the ability of our quadrotor, relying on on-board VIO and visual target localization, to perform aggressive perching maneuvers. The entire control, planning, and target localization pipeline runs on-board the vehicle. In this way, we verify that our system can autonomously perch in  real-world settings. We additionally seek to prove that the AVP strategy provides increased accuracy and precision guaranteeing reliable target interception compared to one-shot planning, especially when increasing the vehicle-target relative distance and for motions with high angular rates (up to $750$~deg/s as demonstrated in Fig.~\ref{fig:angular_rate}).

Our first set of experiments in the real world aims to compare a one-shot planning to AVP. We start by considering the perching problem relying on the Vicon motion capture system for quadrotor localization and employing the camera for target identification and localization. The use of the motion capture system for localization limits the variance between experiments introduced by VIO to obtain a more fair comparison between the one-shot planning and the proposed AVP method. In this way, we can identify the role and benefits or consecutive visual target detection. This experiment is repeated $5$ times for each different target distance.  The average results of the tracking error across all trials are presented in Table~\ref{tab:avp_oneshot}. We define tracking error as the difference between the quadrotor's desired position and the true location of the quadrotor. Additionally, the final two rows $\mathbf{s}_2$ and $\mathbf{s}_3$ of the table quantify the distance of the vehicle with respect to the center of the target once perching is finalized. We notice a substantial improvement to target interception especially once increasing the vehicle-target relative distance. This behavior is demonstrated across several trials in the attached multimedia material. One-shot planning is also unable to intercept a target distant $4.5~\si{m}$ along the $\mathbf{s}_3$. This is because for large distances visual target detection and localization become unreliable and therefore consecutive target interceptions once approaching the target to provide a clear benefit. In a separate table, we also show the target interception error when performing lateral movements in Table~\ref{tab:lateral_error} for the AVP interception. Here, we observe that our target interception remains consistent with and without lateral movements.

\begin{figure}[!b]
    \centering
    \includegraphics[width=\linewidth]{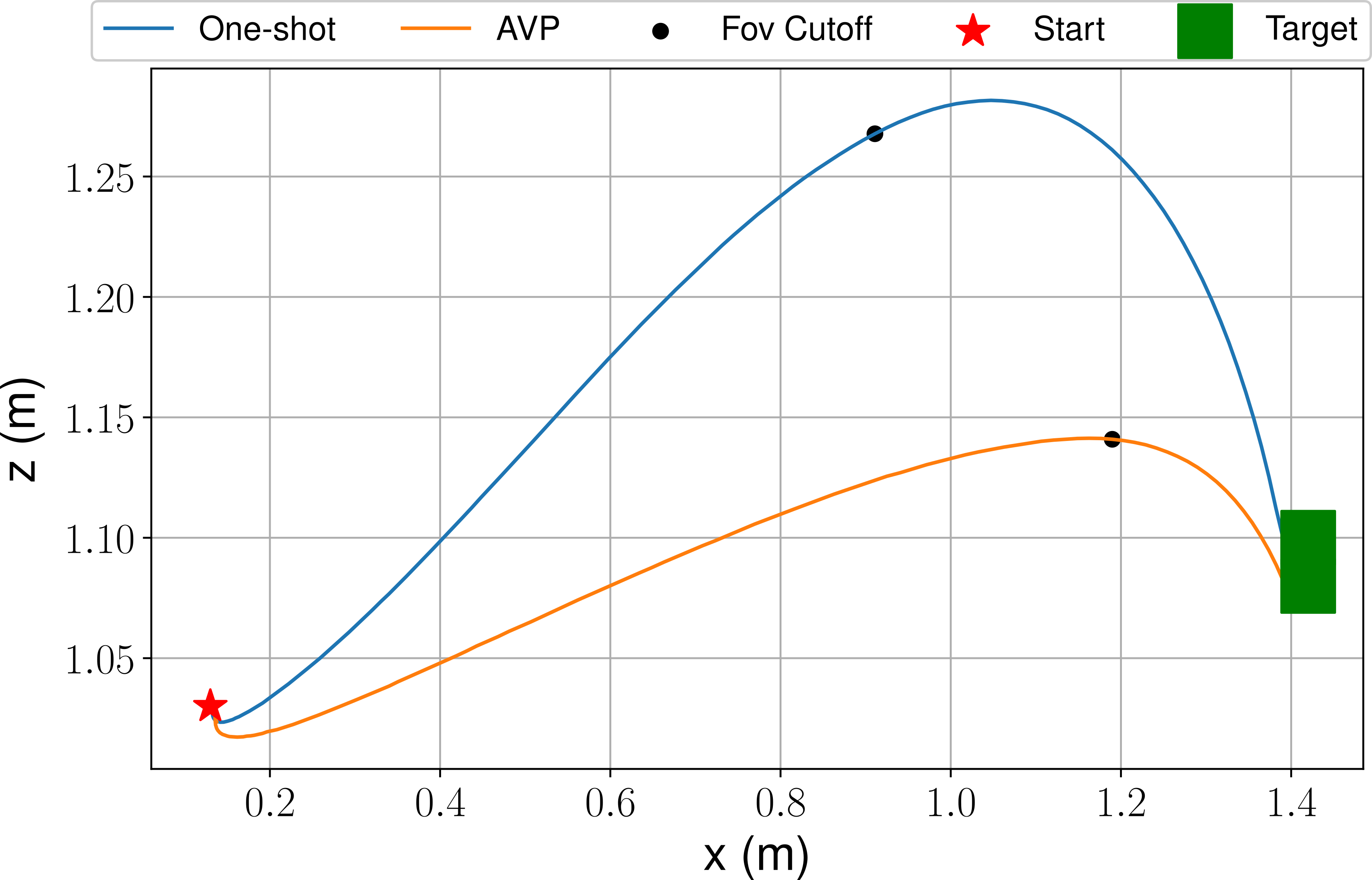}
    \caption{Path comparison between the one-shot and the AVP on the $x$ and $z$ axis.  This experiment was done trying to land on a $90^\circ$ inclined surface with the $\mathbf{v}_{s1} = -2$ and $\mathbf{v}_{s3} = 0.3$. Slight deviations between the end and start points are due to slight difference in AprilTag detection for real world experiments and drift when take off. One-shot perching is done with the Vicon for its path estimation.}
    \label{fig:Path_comp}
\end{figure}

An interesting property that comes out of these trials between one-shot and AVP is that the executed trajectories can be quite different even with accurate state estimation. This is demonstrated in Fig.~\ref{fig:Path_comp} for a path with a $90^\circ$ perching maneuver. We notice that the executed path for the one-shot trajectory case is quite different compared to the AVP trajectory (a similar behavior is experienced in simulation).  Overall in our tests, the trajectory generated by our AVP planner path is shorter.

\begin{table}[!t]
\centering
\caption {Tracking RMSE comparison between AVP and one-shot planning for $90^\circ$ target inclination.  All other axis have a distance of zero when starting. $\mathbf{s}_2$ and $\mathbf{s}_3$ refer to the endpoint error with respect to the target center. - refers to no available data due to unsuccessful perching.}\label{tab:avp_oneshot}
\begin{tabular}{ c c c c c c c}
    \hline
 \rule{0pt}{2ex} Axis&\multicolumn{2}{c}{$1.5$ \si{m}}&\multicolumn{2}{c}{$3.5$ \si{m}}&\multicolumn{2}{c}{$4.5$ \si{m}}\\
  \cline{2-7}\
  \rule{0pt}{2ex}  & One-Shot    & AVP  & One-Shot & AVP & One-Shot & AVP\\\hline
  $x~(\si{m})$&$0.08$  & $0.07$      &  $0.05$& $0.05$ & $0.04$  &  $0.03$ \\
  $y~(\si{m})$&$0.03$   & $0.02$   & $0.01$  &$0.01$  & $0.01$  &  $0.02$ \\
$z~(\si{m})$&$0.04$ & $0.03$   &    $0.03$ & $0.01$   & $0.06$ &  $0.02$ \\
            
    \hline
                 $\mathbf{s}_2~(\si{m})$ & $0.02$  & $0.03$ & $0.05$  & $0.02$ 
            & - & $0.02$  \\
             $\mathbf{s}_3~(\si{m})$& $0.11$   & $0.05$ & $0.15$   & $0.05$ 
            & - & $0.06$  \\\hline

\end{tabular}
\end{table}

\begin{table}[!t]
\begin{center}
\caption{Lateral target interception error using Vicon for vehicle localization and vision for target localization with AVP.
}\label{tab:lateral_error}
\begin{tabular}{c c c c c c}
\hline
  $\mathbf{s}_1$~(\si{m}) &  $\mathbf{s}_2$~(\si{m}) &  $\mathbf{s}_3$~(\si{m}) & & Error $\mathbf{s}_2$~(\si{m}) &  Error $\mathbf{s}_3$~(\si{m})\\
 \cline{1-3}  \cline{5-6}

 $2.5$ & $1$ & $-0.1$ &  &$0.03$ & $0.05$\\
  $2.5$ & $-1$ & $-0.1$ & & $0.03$ & $0.04$\\
  $3.5$ & $0$ & $0$ & & $0.02$ & $0.05$\\
\hline
\end{tabular}
\end{center}
\end{table}

\begin{table}[!b]
\centering
\caption {Estimation RMSE for perching using VIO and AVP. This is the error difference between the motion capture system and the VIO state estimation. \label{tab:tracking_error}} 
\resizebox{\columnwidth}{!}{
\begin{tabularx}{0.53\textwidth}{c c c c c c c c}
    \hline
 \rule{0pt}{2ex} & Axis &\multicolumn{2}{c}{1.5 m}&\multicolumn{2}{c}{3.5 m}&\multicolumn{2}{c}{4.5 m}\\
  \cline{3-8}\
  \rule{0pt}{2ex}  & & One-Shot  & AVP  & One-Shot & AVP & One-Shot & AVP \\\hline
  $60^\circ$&$x~(\si{m})$& $0.03$  & $0.04$ & $0.07$  & $0.07$ & $0.1$ & $0.06$\\
            &$y~(\si{m})$& $0.02$  & $0.02$ & $0.03$  & $0.02$ & $0.03$ & $0.03$\\
            &$z~(\si{m})$& $0.05$  & $0.02$ & $0.04$  & $0.02$ & $0.05$ & $0.09$\\\hline
  $90^\circ$&$x~(\si{m})$& $0.03$  & $0.02$ & $0.04$  & $0.04$ & $0.12$ & $0.1$\\
            &$y~(\si{m})$& $0.02$  & $0.04$ & $0.03$  & $0.05$ & $0.05$ & $0.05$\\
            &$z~(\si{m})$& $0.06$  & $0.03$ & $0.06$  & $0.04$ & $0.08$ & $0.04$ \\\hline
\end{tabularx}}
\vspace{-10pt}
\end{table}
\begin{table}[!t]
\begin{center}
\caption {Target interception perching error at $90^\circ$ surface inclination considering VIO for vehicle localization and vision for target localization.}
\begin{tabular}{ c c c c c c c}
    \hline
 Axis&\multicolumn{2}{c}{$1.5~\si{m}$}&\multicolumn{2}{c}{$3.5~\si{m}$}&\multicolumn{2}{c}{$4.5~\si{m}$}\\
\cline{2-7}\
            & One-Shot   & AVP & One-Shot   & AVP & One-Shot   & AVP\\\hline
            $\mathbf{s}_2~(\si{m})$ &  $0.03$ & $0.05$ & $0.08$    & $0.07$ 
            & --  & $0.07$  \\
            $\mathbf{s}_3~(\si{m})$& $0.10$  & $0.02$ & $0.12$  & $0.06$ 
            & --  & $0.18$ \\\hline
\end{tabular}
 \label{tab:VIO_center_error}
\vspace{-10pt}
\end{center}
\end{table}

\begin{figure*}[!t]
    \centering
     \includegraphics[width=\linewidth,]{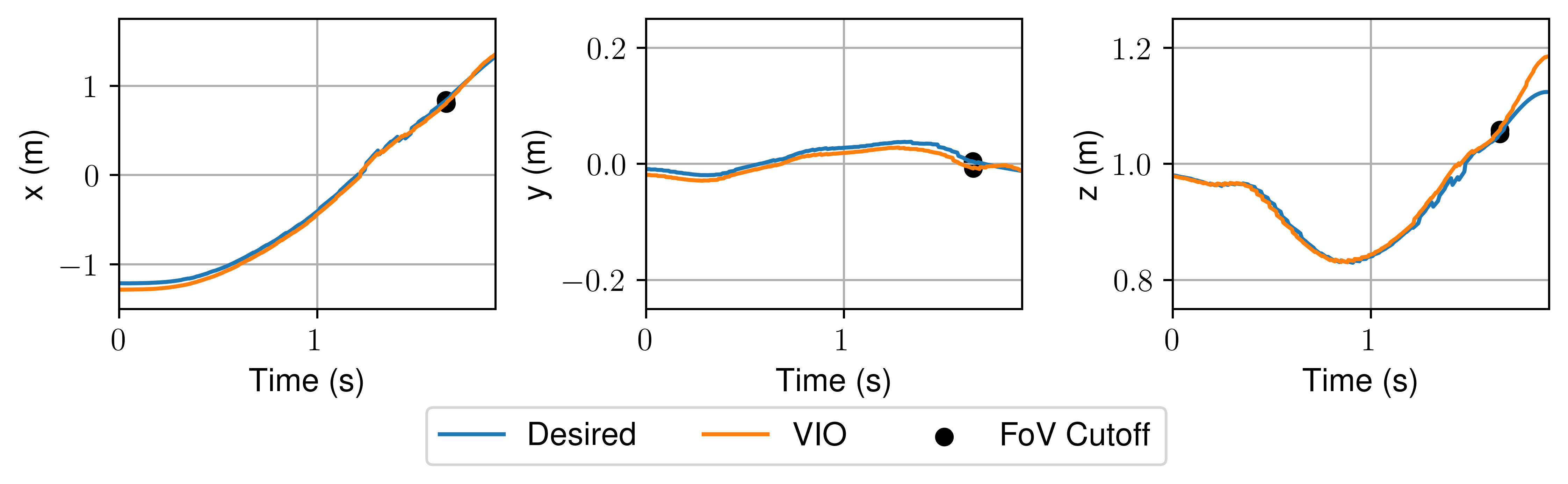}
    \caption{Trajectory tracking and localization visualization for perching on a $90^{\circ}$   surface inclination from a distance of $3~\si{m}$. FoV Cutoff is when the quadrotor loses track of the target.}
    \label{fig:Tracking}
\end{figure*} 


\begin{figure*}[!t]
    \centering
    \includegraphics[width=\linewidth]{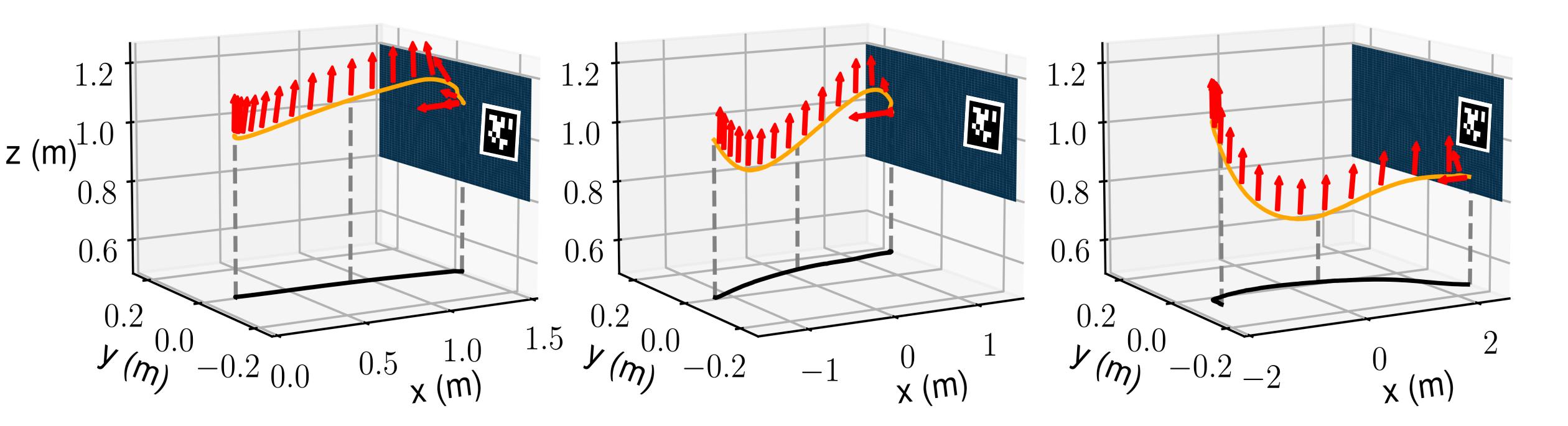}

    \caption{3D path perching trajectory during from various distance, $1.5/3/4.5~\si{m}$ from left to right.  The red arrow is the $\mathbf{b}_3$ axis of the quadrotor. The orange path is the quadrotor's path. The blue board is the target.   The arrows represent the thrust vector. The blue surface represents the target.}
    \label{fig:3d_path}
\vspace{-10pt}
\end{figure*}
\begin{figure*}[!ht]
    \centering
    \includegraphics[width=\linewidth]{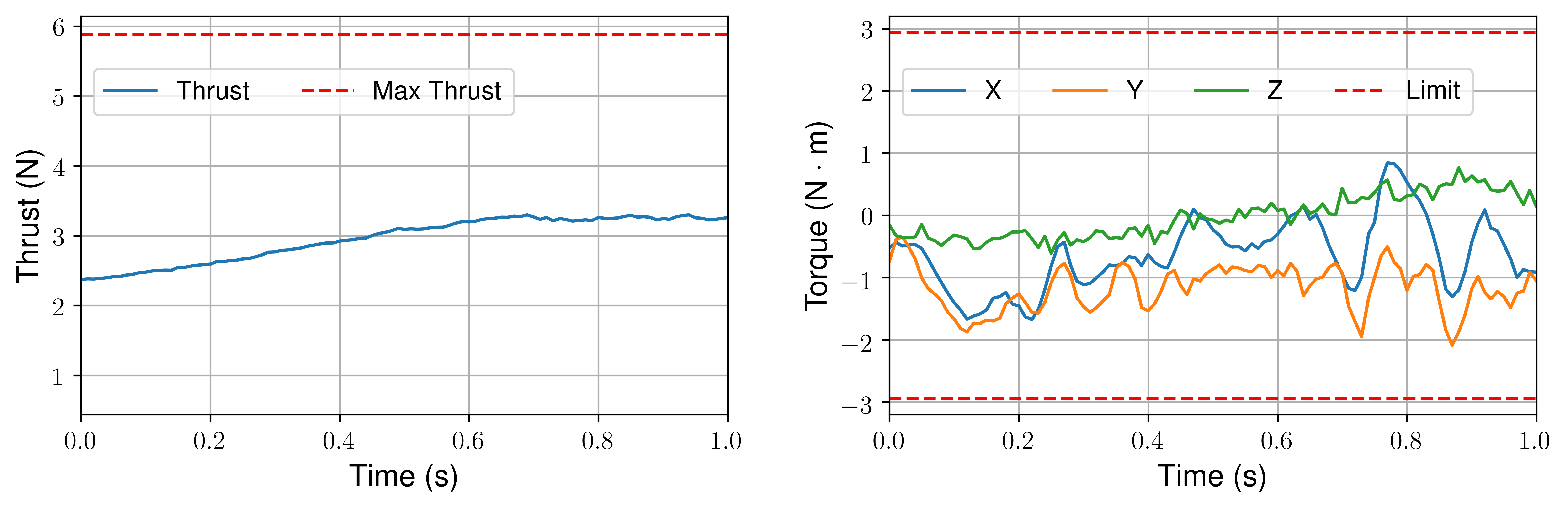}
    \caption{Quadrotor's thrust and moments during perching. Values are calculated directly from motor RPMs readings.  }
    \label{fig:limits}
\end{figure*}

Finally, we perform perching using vision for target and vehicle localization using VIO both running on-board concurrently with planning and control.  A visualization of the estimation error is shown in  Fig.~\ref{fig:Tracking} for a three meter flight at $90^\circ$. We also visualize the paths and thrust  executed by the quadrotor using VIO feedback in Fig.~\ref{fig:3d_path}. The executed path respects the actuators and orientation constraints described in Section~\ref{sec:FoV_const} as shown in Figs.~\ref{fig:3d_path} and~\ref{fig:limits}. The vehicle correctly rotates before target interception and aligns its thrust direction with the target's normal surface. As observed in Fig.~\ref{fig:3d_path}, the further the distance the worse the target interception becomes and the VIO drift also starts to be relevant.  The estimation error is the difference between the Vicon ground truth and the VIO. We quantify the average result of the estimation error in $10$ flight trials in Table~\ref{tab:tracking_error}. We notice that as expected less aggressive maneuvers like perching on  $60^{\circ}$ presents lower tracking and estimation errors compared to more aggressive maneuvers required to perch at $90^{\circ}$. This is mainly due to two factors. First, by increasing the surface inclination, the vehicle will have to perform a larger rotational excursion producing an earlier relaxation of the FoV constraint.  Second, the larger inclination will also increase the difficulty of concurrently guaranteeing a given attitude at a specified location in space due to the system underactuation. Therefore, by increasing the surface inclination, the vehicle has to gain an increased momentum and acceleration to perch. Additionally, we compare the target interception between one-shot and AVP using VIO for localization in Table~\ref{tab:VIO_center_error}. The estimation error is small and similar in both cases due to the limited VIO drifts in  the proposed traveled distances. Similar to our previous experiments with Vicon, one-shot perching is unable to intercept the target placed at $4.5~\si{m}$ distance and on average the AVP performs better than one-shot. We also verify that our trajectory is inside the thrust and moment bounds demonstrated in Fig.~(\ref{fig:limits}). Finally, we compare our success rate of the current interception with our past work over $10$ trials for each surface inclination. For short distances such as $1.5~\si{m}$, we do not see much of a difference from $80\%$ for one-shot and $90\%$ for AVP. For $3.5~\si{m}$, we notice a more substantial difference in success rate $70\%$ for one-shot and $90\%$ for AVP. At further distances such as $4.5~\si{m}$, one-shot is unable to guarantee successful perching. Conversely, AVP guarantees a $60\%$ success rate. Overall, the AVP can reduce and compensate for the effects of inaccurate target localization and control or estimation errors enabling successful and reliable perching compared to the one-shot solution.

\section{Conclusion}
In this paper, we proposed an AVP strategy that exploits in an active fashion mid-flight information such as the vehicle odometry and new target localization to improve the robustness to localization and control errors of our perching task compared to a one-shot approach. We theoretically and experimentally showed how to efficiently formulate and execute this problem as a QP optimization that incorporates actuator and FoV, several perching constraints, and boundary conditions.  The KKT conditions further guarantee the feasibility of the planned maneuver. We experimentally analyze the effects of different boundary conditions on the spatio-temporal target visibility during the perching maneuver and identify the best parameters to maximize it.
The proposed AVP consistently shows superior performances compared to a one-shot planning approach. This algorithm is still lightweight to run  on-board concurrently with the target localization, on-board VIO, and control during flight.

Future works will focus on intercepting a moving target fully exploiting the potential of our planner to adapt in mid-flight. Additionally, it is relevant to improve our controller to incorporate other high speed dynamic effects or disturbances such as drag. We would also like to exploit past perching iterations to further refine the maneuver in a learning-based fashion. Finally, we would also like to explore the possibility to consider a more general object or surface detection algorithm to remove the need for a fiducial marker. This would also facilitate the execution of perching experiments in outdoor settings. 

\bibliographystyle{IEEEtran}
\bibliography{root}

\begin{thebibliography}{10}
\providecommand{\url}[1]{#1}
\csname url@samestyle\endcsname
\providecommand{\newblock}{\relax}
\providecommand{\bibinfo}[2]{#2}
\providecommand{\BIBentrySTDinterwordspacing}{\spaceskip=0pt\relax}
\providecommand{\BIBentryALTinterwordstretchfactor}{4}
\providecommand{\BIBentryALTinterwordspacing}{\spaceskip=\fontdimen2\font plus
\BIBentryALTinterwordstretchfactor\fontdimen3\font minus
  \fontdimen4\font\relax}
\providecommand{\BIBforeignlanguage}[2]{{%
\expandafter\ifx\csname l@#1\endcsname\relax
\typeout{** WARNING: IEEEtran.bst: No hyphenation pattern has been}%
\typeout{** loaded for the language `#1'. Using the pattern for}%
\typeout{** the default language instead.}%
\else
\language=\csname l@#1\endcsname
\fi
#2}}
\providecommand{\BIBdecl}{\relax}
\BIBdecl

\bibitem{MAO_visaul_perch}
J.~Mao, G.~Li, S.~M. Nogar, C.~M. Kroninger, and G.~Loianno, ``Aggressive
  visual perching with quadrotors on inclined surfaces,'' \emph{IEEE/RSJ
  International Conference on Intelligent Robots and Systems (IROS)}, 2021.

\bibitem{Roypoly}
C.~Richter, A.~Bry, and N.~Roy, \emph{Polynomial Trajectory Planning for
  Aggressive Quadrotor Flight in Dense Indoor Environments}.\hskip 1em plus
  0.5em minus 0.4em\relax Cham: Springer International Publishing, 2016, pp.
  649--666.

\bibitem{wang2020alternating}
Z.~Wang, X.~Zhou, C.~Xu, and F.~Gao, ``Alternating minimization based
  trajectory generation for quadrotor aggressive flight,'' \emph{IEEE Robotics
  and Automation Letters (RAL)}, vol.~5, pp. 4836--4843, 2020.

\bibitem{FeiGao_opt_time}
F.~Gao, W.~Wu, J.~Pan, B.~Zhou, and S.~Shen, ``Optimal time allocation for
  quadrotor trajectory generation,'' \emph{IEEE/RSJ International Conference on
  Intelligent Robots and Systems (IROS)}, pp. 4715--4722, 2018.

\bibitem{burke2020generating}
D.~Burke, A.~Chapman, and I.~Shames, ``Generating minimum-snap quadrotor
  trajectories really fast,'' \emph{IEEE/RSJ International Conference on
  Intelligent Robots and Systems (IROS)}, pp. 1487--1492, 2020.

\bibitem{Watterson_SE3}
M.~Watterson, S.~Liu, K.~Sun, T.~Smith, and V.~Kumar, ``Trajectory optimization
  on manifolds with applications to quadrotor systems,'' \emph{The
  International Journal of Robotics Research (IJRR)}, vol.~39, no. 2-3, pp.
  303--320, 2020.

\bibitem{liu2017searchbased}
S.~Liu, K.~Mohta, N.~Atanasov, and V.~Kumar, ``Search-based motion planning for
  aggressive flight in se(3),'' \emph{IEEE Robotics and Automation Letters
  (RAL)}, vol.~PP, 10 2017.

\bibitem{agrawal2021continuoustime}
V.~Agrawal and F.~Dellaert, ``Continuous-time state and dynamics estimation
  using a pseudo-spectral parameterization,'' \emph{IEEE International
  Conference on Robotics and Automation (ICRA)}, pp. 426--432, 2021.

\bibitem{CPC_Philipp}
P.~Foehn, A.~Romero, and D.~Scaramuzza, ``Time-optimal planning for quadrotor
  waypoint flight,'' \emph{Science Robotics}, vol.~6, no.~56, 2021.

\bibitem{Min_energy_traj}
F.~{Morbidi}, R.~{Cano}, and D.~{Lara}, ``Minimum-energy path generation for a
  quadrotor uav,'' \emph{IEEE International Conference on Robotics and
  Automation (ICRA)}, pp. 1492--1498, 2016.

\bibitem{PowerlinePerch}
J.~L.~Paneque, J.~Martinez-de Dios, A.~Ollero, D.~Hanover, S.~Sun, A.~Romero,
  and D.~Scaramuzza, ``Perception-aware perching on powerlines with
  multirotors,'' \emph{IEEE Robotics and Automation Letters}, vol.~7, pp. 1--1,
  04 2022.

\bibitem{acados}
R.~Verschueren, G.~Frison, D.~Kouzoupis, J.~Frey, N.~v. Duijkeren, A.~Zanelli,
  B.~Novoselnik, T.~Albin, R.~Quirynen, and M.~Diehl, ``acados—a modular
  open-source framework for fast embedded optimal control,'' \emph{Mathematical
  Programming Computation}, vol.~14, no.~1, pp. 147--183, 2022.

\bibitem{Thomas_perch_inclined}
J.~Thomas, M.~Pope, G.~Loianno, E.~Hawkes, M.~Estrada, H.~Jiang, M.~Cutkosky,
  and V.~Kumar, ``Aggressive flight for perching on inclined surfaces,''
  \emph{Journal of Mechanisms and Robotics}, vol.~8, 12 2015.

\bibitem{loianno_est}
G.~Loianno, C.~Brunner, G.~Mcgrath, and V.~Kumar, ``Estimation, control, and
  planning for aggressive flight with a small quadrotor with a single camera
  and imu,'' \emph{IEEE Robotics and Automation Letters (RAL)}, vol.~PP, pp.
  1--1, 11 2016.

\bibitem{Chen_acrobat}
Y.~Chen and N.~O. Pérez-Arancibia, ``Controller synthesis and performance
  optimization for aerobatic quadrotor flight,'' \emph{IEEE Transactions on
  Control Systems Technology (TCST)}, vol.~28, no.~6, pp. 2204--2219, 2020.

\bibitem{Lupashin_flip}
S.~Lupashin, A.~Schöllig, M.~Sherback, and R.~D'Andrea, ``A simple learning
  strategy for high-speed quadrocopter multi-flips,'' \emph{IEEE International
  Conference on Robotics and Automation (ICRA)}, pp. 1642--1648, 2010.

\bibitem{Kaufmann_acrobat}
E.~Kaufmann, A.~Loquercio, R.~Ranftl, M.~M{\"{u}}ller, V.~Koltun, and
  D.~Scaramuzza, ``Deep drone acrobatics,'' \emph{Robotics: Science and Systems
  (RSS)}, 2020.

\bibitem{habas2022inverted}
B.~Habas, B.~AlAttar, B.~Davis, J.~W. Langelaan, and B.~Cheng, ``Optimal
  inverted landing in a small aerial robot with varied approach velocities and
  landing gear designs,'' in \emph{2022 International Conference on Robotics
  and Automation (ICRA)}, 2022, pp. 2003--2009.

\bibitem{Thomas_visual_servoing}
J.~{Thomas}, G.~{Loianno}, K.~{Daniilidis}, and V.~{Kumar}, ``Visual servoing
  of quadrotors for perching by hanging from cylindrical objects,'' \emph{IEEE
  Robotics and Automation Letters (RAL)}, vol.~1, no.~1, pp. 57--64, 2016.

\bibitem{Zhang_optimal_traj}
H.~Zhang, B.~Cheng, and J.~Zhao, ``Optimal trajectory generation for
  time-to-contact based aerial robotic perching,'' \emph{Bioinspiration \&
  Biomimetics}, vol.~14, 10 2018.

\bibitem{olson2011tags}
E.~Olson, ``{AprilTag}: A robust and flexible visual fiducial system,''
  \emph{2011 {IEEE} International Conference on Robotics and Automation
  ({ICRA})}, pp. 3400--3407, May 2011.

\bibitem{UsenkoSPC17}
V.~Usenko, L.~Von~Stumberg, A.~Pangercic, and D.~Cremers, ``Real-time
  trajectory replanning for mavs using uniform b-splines and a 3d circular
  buffer,'' \emph{IEEE/RSJ International Conference on Intelligent Robots and
  Systems (IROS)}, pp. 215--222, 2017.

\bibitem{RAPTOR}
B.~Zhou, J.~Pan, F.~Gao, and S.~Shen, ``Raptor: Robust and perception-aware
  trajectory replanning for quadrotor fast flight,'' \emph{IEEE Transactions on
  Robotics (T-RO)}, 2021.

\bibitem{Penin2018}
B.~Penin, P.~R. Giordano, and F.~Chaumette, ``Vision-based reactive planning
  for aggressive target tracking while avoiding collisions and occlusions,''
  \emph{IEEE Robotics and Automation Letters}, vol.~3, no.~4, pp. 3725--3732,
  2018.

\bibitem{Falanga_vision_perching}
D.~Falanga, A.~Zanchettin, A.~Simovic, J.~Delmerico, and D.~Scaramuzza,
  ``Vision-based autonomous quadrotor landing on a moving platform,''
  \emph{IEEE International Symposium on Safety, Security and Rescue Robotics
  (SSRR)}, pp. 200--207, 2017.

\bibitem{ji2022real}
J.~Ji, T.~Yang, C.~Xu, and F.~Gao, ``Real-time trajectory planning for aerial
  perching,'' in \emph{IEEE/RSJ International Conference on Intelligent Robots
  and Systems (IROS)}, 2022, pp. 10\,516--10\,522.

\bibitem{Chi2014}
W.~{Chi}, K.~H. {Low}, K.~H. {Hoon}, and J.~{Tang}, ``An optimized perching
  mechanism for autonomous perching with a quadrotor,'' \emph{IEEE
  International Conference on Robotics and Automation (ICRA)}, pp. 3109--3115,
  2014.

\bibitem{AGRASP}
K.~M. Popek, M.~S. Johannes, K.~C. Wolfe, R.~A. Hegeman, J.~M. Hatch, J.~L.
  Moore, K.~D. Katyal, B.~Y. Yeh, and R.~J. Bamberger, ``Autonomous grasping
  robotic aerial system for perching (agrasp),'' \emph{IEEE/RSJ International
  Conference on Intelligent Robots and Systems (IROS)}, pp. 1--9, 2018.

\bibitem{roderick2021bird}
W.~Roderick, M.~Cutkosky, and D.~Lentink, ``Bird-inspired dynamic grasping and
  perching in arboreal environments,'' \emph{Science Robotics}, vol.~6, no.~61,
  p. eabj7562, 2021.

\bibitem{Kalantari2015}
A.~{Kalantari}, K.~{Mahajan}, D.~{Ruffatto}, and M.~{Spenko}, ``Autonomous
  perching and take-off on vertical walls for a quadrotor micro air vehicle,''
  \emph{IEEE International Conference on Robotics and Automation (ICRA)}, pp.
  4669--4674, 2015.

\bibitem{Hawkes2013}
E.~W. {Hawkes}, D.~L. {Christensen}, E.~V. {Eason}, M.~A. {Estrada},
  M.~{Heverly}, E.~{Hilgemann}, H.~{Jiang}, M.~T. {Pope}, A.~{Parness}, and
  M.~R. {Cutkosky}, ``Dynamic surface grasping with directional adhesion,''
  \emph{IEEE/RSJ International Conference on Intelligent Robots and Systems
  (IROS)}, pp. 5487--5493, 2013.

\bibitem{Daler2013}
L.~{Daler}, A.~{Klaptocz}, A.~{Briod}, M.~{Sitti}, and D.~{Floreano}, ``A
  perching mechanism for flying robots using a fibre-based adhesive,''
  \emph{IEEE International Conference on Robotics and Automation (ICRA)}, pp.
  4433--4438, 2013.

\bibitem{Kessens2016}
C.~C. {Kessens}, J.~{Thomas}, J.~P. {Desai}, and V.~{Kumar}, ``Versatile aerial
  grasping using self-sealing suction,'' \emph{IEEE International Conference on
  Robotics and Automation (ICRA)}, pp. 3249--3254, 2016.

\bibitem{Backus_gripper_arm}
S.~Backus, J.~Izraelevitz, J.~Quan, R.~Jitosho, E.~Slavick, and A.~Kalantari,
  ``Design and testing of an ultra-light weight perching system for sloped or
  vertical rough surfaces on mars,'' \emph{IEEE Aerospace Conference}, pp.
  1--12, 2020.

\bibitem{wopereis_arm}
H.~W. Wopereis, T.~D. van~der Molen, T.~H. Post, S.~Stramigioli, and
  M.~Fumagalli, ``Mechanism for perching on smooth surfaces using aerial
  impacts,'' \emph{IEEE International Symposium on Safety, Security, and Rescue
  Robotics (SSRR)}, pp. 154--159, 2016.

\bibitem{Park_ceiling}
S.~Park, D.~S. Drew, S.~Follmer, and J.~Rivas-Davila, ``Lightweight high
  voltage generator for untethered electroadhesive perching of micro air
  vehicles,'' \emph{IEEE Robotics and Automation Letters (RAL)}, vol.~5, no.~3,
  pp. 4485--4492, 2020.

\bibitem{Pope_SCAMP}
M.~T. Pope, C.~W. Kimes, H.~Jiang, E.~W. Hawkes, M.~A. Estrada, C.~F. Kerst,
  W.~R.~T. Roderick, A.~K. Han, D.~L. Christensen, and M.~R. Cutkosky, ``A
  multimodal robot for perching and climbing on vertical outdoor surfaces,''
  \emph{IEEE Transactions on Robotics (T-RO)}, vol.~33, no.~1, pp. 38--48,
  2017.

\bibitem{Lussier2011}
A.~L. Desbiens, A.~T. Asbeck, and M.~R. Cutkosky, ``Landing, perching and
  taking off from vertical surfaces,'' \emph{The International Journal of
  Robotics Research (IJRR)}, vol.~30, no.~3, pp. 355--370, 2011.

\bibitem{Moore_2014}
J.~Moore, R.~Cory, and R.~Tedrake, ``Robust post-stall perching with a simple
  fixed-wing glider using lqr-trees,'' \emph{Bioinspiration \& Biomimetics},
  vol.~9, p. 025013, 05 2014.

\bibitem{Mehanovic_fixed_wing}
D.~Mehanovic, J.~Bass, T.~Courteau, D.~Rancourt, and A.~L. Desbiens,
  ``Autonomous thrust-assisted perching of a fixed-wing uav on vertical
  surfaces,'' \emph{Conference on Biomimetic and Biohybrid Systems}, pp.
  302--314, 2017.

\bibitem{LoiannoRAL2017}
G.~Loianno, C.~Brunner, G.~McGrath, and V.~Kumar, ``Estimation, control, and
  planning for aggressive flight with a small quadrotor with a single camera
  and imu,'' \emph{IEEE Robotics and Automation Letters (RAL)}, vol.~2, no.~2,
  pp. 404--411, April 2017.

\bibitem{McLamrock}
T.~Lee, M.~Leok, and N.~H. McClamroch, ``{Nonlinear Robust Tracking Control of
  a Quadrotor UAV on SE(3)},'' \emph{Asian Journal of Control}, vol.~15, no.~2,
  pp. 391--408, 2013.

\bibitem{Loianno_IROS2015}
G.~Loianno, Y.~Mulgaonkar, C.~Brunner, D.~Ahuja, A.~Ramanandan, M.~Chari,
  S.~Diaz, and V.~Kumar, ``Smartphones power flying robots,'' \emph{IEEE/RSJ
  International Conference on Intelligent Robots and Systems (IROS)}, pp.
  1256--1263, Sept 2015.

\bibitem{PCMPC}
G.~Li, A.~Tunchez, and G.~Loianno, ``Pcmpc: Perception-constrained model
  predictive control for quadrotors with suspended loads using a single camera
  and imu,'' \emph{IEEE International Conference on Robotics and Automation
  (ICRA)}, 2021.

\bibitem{sturm2009}
C.~Sturm, ``Collected works of charles françois sturm,'' \emph{Inst. France
  Sc. Math. Phys.}, vol.~6, pp. 345--390, 01 2009.

\bibitem{song2020flightmare}
Y.~Song, S.~Naji, E.~Kaufmann, A.~Loquercio, and D.~Scaramuzza, ``Flightmare: A
  flexible quadrotor simulator,'' in \emph{Conference on Robot Learning}, 2020.

\bibitem{OOQP}
E.~M. Gertz and S.~J. Wright, ``Object-oriented software for quadratic
  programming,'' \emph{ACM Trans. Math. Softw.}, vol.~29, no.~1, p. 58–81,
  Mar. 2003.

\bibitem{SQP}
D.~Kraft, \emph{A Software Package for Sequential Quadratic Programming}, ser.
  Deutsche Forschungs- und Versuchsanstalt f{\"u}r Luft- und Raumfahrt
  K{\"o}ln: Forschungsbericht.\hskip 1em plus 0.5em minus 0.4em\relax Wiss.
  Berichtswesen d. DFVLR, 1988.

\bibitem{Loianno_smartphone}
G.~Loianno, Y.~Mulgaonkar, C.~Brunner, D.~Ahuja, A.~Ramanandan, M.~Chari,
  S.~Diaz, and V.~Kumar, ``A swarm of flying smartphones,'' \emph{IEEE/RSJ
  International Conference on Intelligent Robots and Systems (IROS)}, pp.
  1681--1688, 10 2016.

\bibitem{Mellinger2011}
D.~Mellinger and V.~Kumar, ``{Minimum Snap Trajectory Generation and Control
  for Quadrotors},'' \emph{IEEE International Conference on Robotics and
  Automation (ICRA)}, pp. 2520--2525, 2011.

\end{thebibliography}
\begin{IEEEbiography}[{\includegraphics[width=1.1in,height=1.1in,clip]{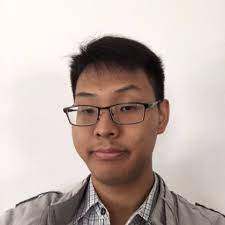}}]{Jeffrey Mao} is a Ph.D. at NYU under the guidance of Prof. Loianno at the Agile Robotics and Perception Lab. His research interest include aerial robotics, trajectory planning, navigation, gaussian processes, and perception. He received his MS at New York University (NYU), USA in Electrical and Computer Engineering in 2019. Prior to joining NYU, he worked at Intel on wearable embedded system devices.
\end{IEEEbiography}
 
\begin{IEEEbiography}[{\includegraphics[width=1.1in,height=1.1in,clip]{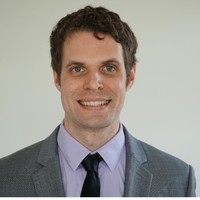}}]{Stephen Nogar} is currently an Autonomy Team Lead at DEVCOM Army Research Laboratory.
 He received a Ph.D. focused in Aerospace Engineering from Ohio State University, USA in 2015. His research experience includes dynamics, control, perception and multi-agent teaming for autonomous systems. 
\end{IEEEbiography}

\begin{IEEEbiography}[{\includegraphics[width=1.1in,height=1.1in]{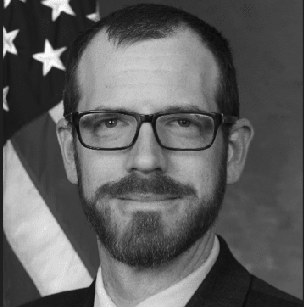}}]{Christopher M. Kroninger } is a Program Manager with the U.S. Army DEVCOM Army Research Laboratory. He received an MS in Aerospace Engineering from the Pennsylvania State University in 2008. He  His research focus is single and multi-agent autonomy to include heterogeneous team control, agent-task assignment, and vision-based navigation.\end{IEEEbiography}

\begin{IEEEbiography}[{\includegraphics[width=1.1in,height=1.1in,clip]{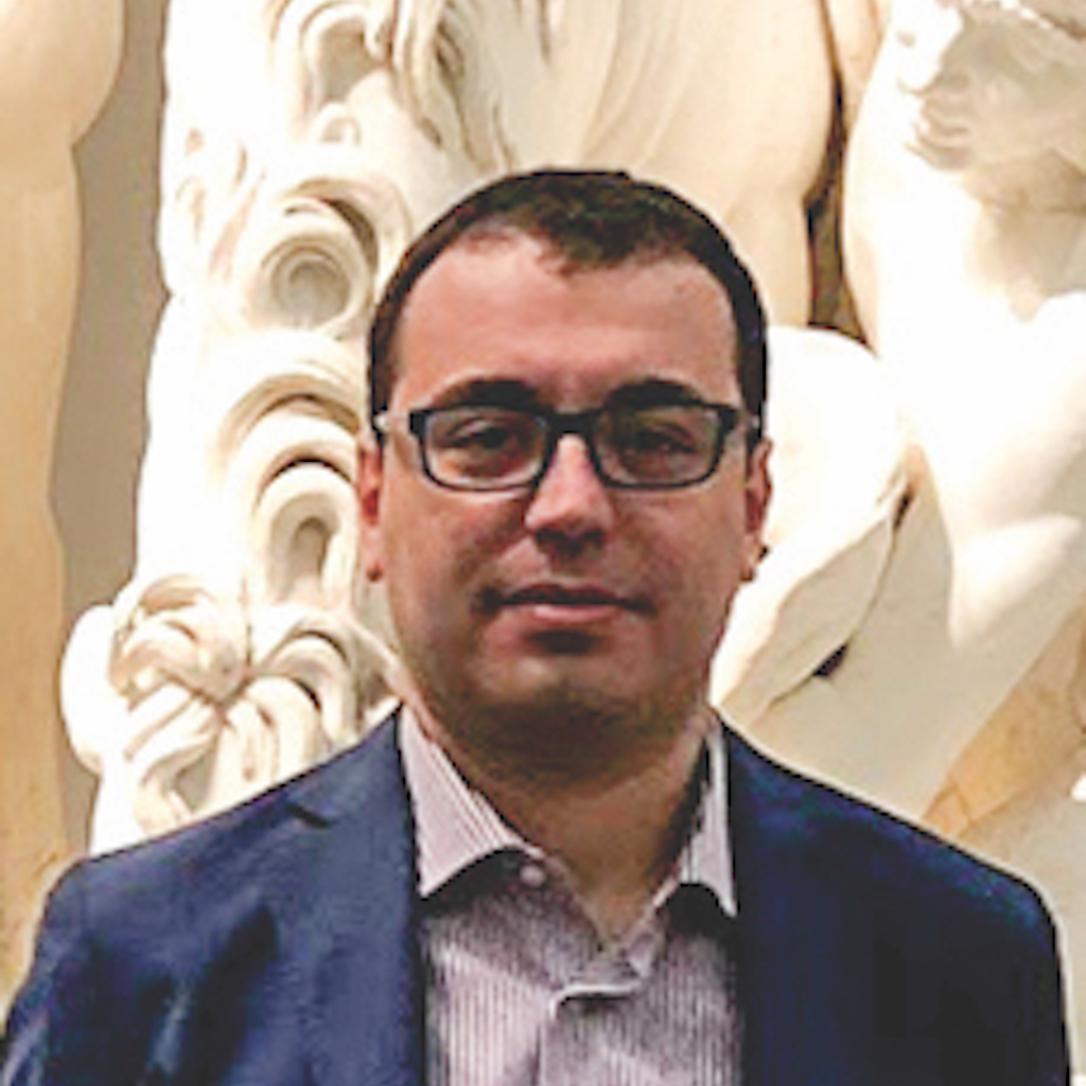}}]{Giuseppe Loianno} is an assistant professor at the New York University, USA and director of the Agile Robotics and Perception Lab (\url{https://wp.nyu.edu/arpl/}) working on autonomous robots. He received a Ph.D. in robotics from University of Naples "Federico II", Italy in 2014. Prior joining NYU, he was post-doctoral researcher, research scientist and team leader at the GRASP Lab at the University of Pennsylvania in Philadelphia, USA. Dr. Loianno has published more than 70 conference papers, journal papers, and book chapters. His research interests include perception, learning, and control for autonomous robots. He received the NSF CAREER Award in 2022 and DARPA Young Faculty Award in 2022. He is recipient of the IROS Toshio Fukuda Young Professional Award in 2022, Conference Editorial Board Best Associate Editor Award at ICRA 2022, Best Reviewer Award at ICRA 2016, and he was selected as Rising Star in AI from KAUST in 2023. He is also currently the co-chair of the IEEE RAS Technical Committee on Aerial Robotics and Unmanned Aerial Vehicles. He was the the general chair of the IEEE International Symposium on Safety, Security and Rescue Robotics (SSRR) in 2021 as well as program chair in 2019, 2020, and 2022. His work has been featured in a large number of renowned international news and magazines.\end{IEEEbiography}
\section*{Appendix}~\label{sec:appendix}
\vspace{-10pt}
\subsection{Sturm's Theorem}
Sturm's theorem~\cite{sturm2009} states that the number of roots for a polynomial, $H\left(t\right)$ in an interval $\left[t_0,t_f\right]$ is equal to the difference in sign changes of the Sturm's sequence, eq.~(\ref{eqn:sturm_seq}), between $S\left(t_0\right)$ and $S\left(t_f\right)$  
\begin{equation}\label{eqn:sturm_seq}
    S\left(t\right) =  \begin{cases}
       S_0\left(t\right) =H\left(t\right) \\
       S_1\left(t\right) =\dot{H}\left(t\right)\\
        S_{i+1}\left(t\right)=  -Rm\left(S_{i-1},S_i\right) \\
        \vdots\\
        S_{N}\left(t\right) = -Rm\left(S_{N-2},S_{N-1}\right) \in \mathbb{R}
     \end{cases},
\end{equation}
where $Rm(S_{i-1}, S_i)$ gives the algebraic remainder of $\displaystyle\frac{S_{i-1}}{S_{i}}$.
To evaluate the number of roots between $t \in \left[0,2\right]$ for $H\left(t\right) = t^4+t^3-t-1$. First calculate the Sturm sequence,
\begin{equation}\label{eqn:sturm_seq_eval}
    S\left(t\right) =  \begin{cases}
       S_0\left(t\right) =t^4+t^3-t-1 \\
       S_1\left(t\right) =4t^3+3t^2-1 \\         S_2\left(t\right) =0.1875t^2+0.75t+0.9375 \\ 
        S_3\left(t\right) =-32t-64 \\        S_4\left(t\right) =-0.1875 \\ 
     \end{cases}.
\end{equation}
We evaluate this sequence's signs at $t=0$ as $\left[-,-,+,-,-\right]$. This sequence has $2$ sign changes. Next, we calculate the sequence at $t=2$ as  $\left[+,+,+,-,-\right]$. There is $1$ sign change. Subtracting the number of sign changes at $t=0$ from $t=1$, we find 1 root for $H(t)$ in the domain $\left[0,2\right]$. 

\textit{Proof.}
To prove that the Sturm's theorem guarantees the values are in the bound we leverage the intermediate value theorem. The intermediate value theorem states that given a continuous function $H\left(t\right)$ whose domain contains the values $\left[t_0,t_f\right]$ then $\forall \hspace{10pt} i \in \left[H(t_0), H(t_f)\right]$ there must exist a corresponding $t_i\in\left[t_0,t_f\right]$ such that $i = H(t_i)$. Since our trajectory is continuous, this theorem holds in our case. Now let's prove that our algorithm works by contradiction. Assume there exists a $t_i\in\left[t_0,t_1\right]$ such that $H\left(t_i\right) > b$ where $b$ is the global bound, and all conditions of Algorithm 1, $H\left(t_0\right) < b$, $H\left(t_f\right) < b$, and $H\left(t_i\right)-b \neq 0~\forall~t_i\in \left[t_0,t_f\right]$ are true.

If this is the case, then we can apply the intermediate value theorem and construct a domain $\left[t_0, t_i\right]$ and a range $\left[H\left(t_0\right), H\left(t_i\right)\right]$.  We know that $H\left(t_0\right) < b$ and $H\left(t_i\right) > b$, then $b \in \left[H\left(t_0\right), H\left(t_i\right)\right]$. Therefore, based on the intermediate value theorem, there must exist a $t_j \in \left[t_0,t_i\right]$ such that $H\left(t_j\right)=b$. However, we see $H\left(t_j\right)=b$ is a contradiction with respect to the condition $H\left(t_j\right)-b \neq 0~\forall~t_j\in \left[t_0,t_f\right]$. As $\left[t_0,t_i\right] \subset \left[t_0,t_f\right]$ by construction, this condition also holds true for all $t_j \in \left[t_0,t_i\right]$. Since this is a contradiction there can exist no such number $t_i \in \left[t_0,t_1\right]$ such that $H\left(t_i\right) > b$ if our algorithm returns true.$\blacksquare$

\subsection{Flat Output Polynomial bounds}\label{sec:poly_bounds}
In this section, we  derive upper bounds for the thrust rate, angular velocity, angular accelerations and moments as functions of the flat outputs represented with polynomials. These condition can eventually be added and complement additional bounds we described in Section~\ref{sec:act_constraints} especially if the motor low-level control setup is not reactive enough. The boundaries can be verified using the Sturms on the derived polynomial bounds and implemented with minimal modifications. The following boundaries on thrust rate and moments are functions of the third and fourth order derivatives of the trajectory, namely the jerk and snap. Therefore, since the third or fourth order derivative norm is already being minimized as our cost, adding additional time to our quadrotor will enable to obtain a feasible trajectory since we are indirectly minimizing angular velocity, thrust rate, and moments by reducing the quadrotor jerk.

\subsubsection{Thrust Rate}
In our previous work~\cite{Loianno_smartphone}, we showed that the thrust is a function of the flat outputs. Let $m$ be the mass of the vehicle, $\mathbf{j}$ the jerk which represents the third order derivative of the positional trajectory, the rotation of the quadrotor defined by eq.~(\ref{eq:R_des}) convention as Z($\psi$)-Y($\theta$) - X($\phi$), and  $\dot\tau$ the thrust rate. It holds that
\begin{equation}
    m\mathbf{b}_3 \cdot \mathbf{j} = \dot\tau
\end{equation}
We can then construct an upper bound of the dot product using the geometric definition 
\begin{equation}
    m\mathbf{b}_3 \cdot \mathbf{j} = m(\left|\left|\mathbf{b}_3\right|\right|)(\left|\left|\mathbf{j}\right|\right|)\cos \theta \leq m(\left|\left|\mathbf{b}_3\right|\right|)(\left|\left|\mathbf{j}\right|\right|),
\end{equation}
where $\theta$ is the angle between $\mathbf{b}_3$ and $\mathbf{j}$. By construction $\left|\left|\mathbf{b}_3\right|\right|=1$, therefore, we can create an upper bounded polynomial such that
\begin{equation}\label{eqn:thrust_rate_bound}
     \dot\tau^2 \leq \left(m\left|\left|\mathbf{j}\right|\right|\right)^2 \leq \dot\tau_{max}^2.
\end{equation}
We choose to use the squared norm to avoid the square root in the norm function. The function $(m\left|\left|\mathbf{j}\right|\right)^2$ remains a polynomial as it is constructed solely through derivatives and multiplication which means the trajectory still remains in the group of polynomial functions.

\subsubsection{Angular Velocity}\label{sec:angular velocity}
Angular velocity is constructed from our formulation of the rotation matrix defined by eq.~(\ref{eq:R_des}). We consider our angular velocity vector components as $\mathbf{\Omega} = [\omega_1~ \omega_2~\omega_3]^\top$

\begin{equation}\label{eqn:w12b12_def}
     \begin{bmatrix}
    \omega_1\\
    \omega_2
\end{bmatrix} = \frac{m}{\tau}\begin{bmatrix}
-\mathbf{b}_2^T\\
\mathbf{b}_1^T
\end{bmatrix} \mathbf{j}.
\end{equation}
By construction $\mathbf{b}_1$ and $\mathbf{b}_2$ are orthogonal. Therefore, we can form the following inequality based on the minimum thrust $\tau$
\begin{equation}\label{eqn:w12_bound}
\omega_1^2+\omega_2^2 \leq 2\frac{m}{\tau}\left|\left|\mathbf{j}\right|\right|^2.
\end{equation}

While this eq.~(\ref{eqn:w12_bound}) may not be a polynomial an upper bounded polynomial can be found by simply rounding the function up as both thrust and norm of jerk are positive thus allowing us to apply  Sturm's Theorem.

The third component of the angular velocity requires us to declare the Euler angle convention. We can derive the third component through the flat output as
\begin{equation}\label{eqn:omega_3orig_def}
    \omega_3 = \frac{\cos \theta}{\cos \phi} \dot\psi -\omega_2 \tan \phi.
\end{equation}
Eq.~(\ref{eqn:omega_3orig_def}) has a singularity when the roll or $\phi = \pi/2$ unlike the formulation for the Euler angles in our previous work \cite{Loianno_smartphone} which has the singularity on the pitch at the same value. Generally, as in the presented case, perching is a maneuver which relies on pitching more than rolling. This is because the front camera is placed along the the forward motion direction of the quadrotor to facilitate to maintain line of sight of the target. As a result, the quadrotor will be primarily pitching during the interception maneuver. An interception through rolling would make it impossible for a front camera to maintain the target in the FoV. As such, we set an assumption that the roll is bounded to the following angles
\begin{equation}\label{eqn:roll_limit}
    -\frac{\pi}{4} \leq \phi \leq \frac{\pi}{4}.
\end{equation}
Following this assumption, we can construct this bound on the angular velocity
\begin{equation}\label{eqn:w3_2_ref}
    \omega_3^2 \leq \frac{1}{2}\left(\dot\psi\cos\theta - \omega_2\sin \phi\right)^2.
\end{equation}
The expression $\dot\psi\cos\theta - \omega_2\sin \phi$ is upper bounded with the following eq.~(\ref{eqn:w3_def})
\begin{equation}\label{eqn:w3_def}
    \left(\dot\psi\cos\theta - \omega_2\sin \phi\right)^2 \leq \dot\psi^2+\omega_2^2+2|\dot\psi\omega_2|.
\end{equation}
From this result, we can construct an upper bound on $w_3^2$ from eq.~(\ref{eqn:w3_def}) by substituting the absolute value function with $|x| < 0.1x^2+2.5$ this bound was picked because of its implementation simplicity and being fairly similar near the end of the angular velocity range
\begin{equation}\label{eqn:w3_bound}
\begin{aligned}
     \omega_3^2   &\leq \frac{1}{2}\left(\dot\psi^2 + \omega_2^2\right) +|\dot\psi\omega_2|, \\
     \omega_3^2   &\leq \frac{1}{2} \left(\dot\psi^2 +\frac{m}{\tau}\left|\left|\mathbf{j}\right|\right|^2+0.2\frac{m}{\tau}\left|\left|\mathbf{j}\right|\right|^2\dot\psi^2 \right) +2.5. \\
\end{aligned}
\end{equation}
A closer bound on absolute value can be solved with a polynomial at the cost of an increased complexity. Combining the eqs.~(\ref{eqn:w12_bound}) and~(\ref{eqn:w3_bound}), we can construct our upper bound as
\begin{equation}~\label{eqn:omega_bound}
  \omega_1^2+\omega_2^2+\omega_3^2 \leq \frac{5m}{2\tau}\left|\left|\mathbf{j}\right|\right|^2 +\frac{1}{2} \dot \psi^2+\frac{m}{10\tau}\left|\left|\mathbf{j}\right|\right|^2\dot\psi^2+2.5.
\end{equation}
The above upper bound function can be constructed by multiplication and derivatives of a polynomial with other polynomials and scalars. As a result, this means the function also can be perfectly represented by a polynomial. This means that the Sturm's algorithm can be applied to this formulation. 

\subsubsection{Angular acceleration}
The angular acceleration can be derived in a similar manner by taking another derivative.
From our eq.~(\ref{eqn:model}), we can take two derivatives on the translational and given  dynamics to form the following equality and substituting $\frac{d^4\mathbf{x}}{dt^4} = \mathbf{s}$
\begin{equation}\label{sec:second order tranlsational}
    m\mathbf{s} = 2\dot{\tau}\mathbf{R}\hat{\mathbf{\Omega}}\mathbf{e}_3 + \tau\mathbf{R}\hat{\mathbf{\Omega}}^2\mathbf{e}_3  +\tau\mathbf{R}\hat{\dot{\mathbf{\Omega}}}\mathbf{e}_3  + \ddot{\tau}\mathbf{R}\mathbf{e}_3.
\end{equation}
We can formulate the first two components of the angular acceleration, $\dot{\omega}_1$ and $\dot{\omega}_2$, by solving for $\hat{\dot{\mathbf{\Omega}}}\mathbf{e}_3$ 
\begin{equation}
\begin{split}
     &\begin{bmatrix}
    \dot{\omega}_1\\
    \dot{\omega}_2
\end{bmatrix} = \frac{m}{\tau}\begin{bmatrix}
-\mathbf{b}_2^T\\
\mathbf{b}_1^T
\end{bmatrix}\mathbf{s}  +\\ &\frac{2\dot{\tau}m}{\tau^2}\begin{bmatrix}
\mathbf{b}_2^T\\
-\mathbf{b}_1^T
\end{bmatrix}\mathbf{j}+\omega_3\frac{m}{\tau}\begin{bmatrix}
-\mathbf{b}_1^T\\
\mathbf{b}_2^T
\end{bmatrix} \mathbf{j}.
\end{split}
\end{equation}
We can construct an upper bound in the form of a polynomial as
\begin{equation}\label{eqn:w12_dot}
    \dot{\omega}_1^2+
    \dot{\omega}_2^2
 \leq 2\frac{m}{\tau}\left|\left|\mathbf{s}  \right|\right|^2 +4\frac{m\dot\tau}{\tau^2}\left|\left|\mathbf{j}\right|\right|^2+\frac{m\omega_3}{\tau}\left|\left|\mathbf{j}\right|\right|^2.
\end{equation}
Moving forward, we can substitute $\omega_3$ with a variation of eq.~(\ref{eqn:w3_2_ref}) to form the following inequality. 
\begin{equation}
\begin{split}
    &\dot{\omega}_1^2+
    \dot{\omega}_2^2
 \leq 2\frac{m}{\tau}\left|\left|\mathbf{s}\right|\right|^2 +\\
 &4\frac{m\dot\tau}{\tau^2}\left|\left|\mathbf{j}\right|\right|^2+\frac{m}{\tau}\left|\left|\mathbf{j}\right|\right|^2\left(|\dot\psi|+
 \frac{m}{\tau}\left|\left|\mathbf{j}\right|\right|\right).
\end{split}
\end{equation}
While absolute value is not a polynomial, it can be substituted with a polynomial bound that follows it fairly similarly.
We can further derive $\dot{\omega}_3$ from the relationship of our rotation matrix construction described in eq.~(\ref{eq:R_des}) considering a Z-Y-X convention
\begin{equation}\label{eqn:omega_3dotorig_def}
\begin{split}
    &\dot{\omega}_3 = \omega_3\tan \phi+\\ &\frac{1}{\cos \phi}\left(\ddot{\psi}\cos\theta -\dot\psi\dot\theta\sin\theta-\dot{\omega}_2\sin\phi-\omega_2\dot{\phi}\cos\phi\right) .
    \end{split}
\end{equation}
A simplified version factoring out the euler angle derivatives for pitch $\dot\theta$ and roll $\dot\phi$ can be generated if we also set the same roll limit described in eq.~(\ref{eqn:roll_limit})
\begin{equation}\label{eqn:omega_3dot_bound}
    \dot{\omega}_3 \leq \frac{\sqrt{2}}{2}\left(\left|\ddot \psi\right| + \sqrt{\dot\omega_2^2 +\left(\omega_1\omega_2\right)^2}+\dot\psi^2 +\left|\dot\psi\omega_2\right|\right) +\left|\omega_3\right|
\end{equation}
We can let eq.~(\ref{eqn:w12_dot}) be represented as $F(\dot\psi,\mathbf{a},\mathbf{j},\mathbf{s})$ and  eq.~(\ref{eqn:omega_3dot_bound}) be represented as $G(\ddot\psi,\dot\psi,\mathbf{a},\mathbf{j},\mathbf{s})$ and substituting in respectively eqs.~(\ref{eqn:w12b12_def}) and eq.~(\ref{eqn:omega_3orig_def}) we obtain 
\begin{equation}\label{eqn:omega_bound_dot}
    \dot{\omega}_1^2+
    \dot{\omega}_2^2+\dot{\omega}_3^2
 \leq F\left(\psi,\dot\psi,\mathbf{a},\mathbf{j},\mathbf{s}\right) + G\left(\psi,\dot\psi,\ddot\psi,\mathbf{a},\mathbf{j},\mathbf{s}\right)^2.
\end{equation}

\subsubsection{Moments}~\label{sec:moment}
We can derive an upper bound on the moments based on eq.~(\ref{eqn:model}). By considering a linear relationship between force and moments with motor speeds~\cite{LoiannoRAL2017}, by deriving a boundary on the moment then it is simple to derive a boundary on the motor speeds. Without loss of generality, to make the derivation simpler, we assume the components are solely diagonal for our inertial matrix. This is also a common assumption on most quadrotors. There exists a more complex version for a non-diagonal inertial matrix. Let the matrix components of the inertial matrix be derived as 
\begin{equation}
     \mathbf{J} = \begin{bmatrix}
         I_{xx}^2 & 0 & 0\\
         0 &  I_{yy}^2& 0\\
         0 & 0 &  I_{zz}^2
     \end{bmatrix}.
\end{equation}
First, let $\mathbf{\dot\Omega} \leq H(\ddot\psi,\dot\psi,\mathbf{j},\mathbf{s})$ where $H(\ddot\psi,\dot\psi,\mathbf{j},\mathbf{s})$ is the inequality derived in the previous section in eq.~(\ref{eqn:omega_bound}), it holds that
\begin{equation}\label{eqn:moment_det}
    \mathbf{M} = \mathbf{J}\mathbf{\dot\Omega} + \mathbf{\Omega} \times \mathbf{J}\mathbf{\Omega}.
\end{equation}
Leveraging the definition of our inertial matrix, we can formulate this cross product as a product of a constant matrix and a vector as
\begin{equation}
\mathbf{\Omega} \times \mathbf{J}\mathbf{\Omega} = \mathbf{I}\begin{bmatrix}
 \omega_2\omega_3\\
    \omega_1\omega_3\\
    \omega_1\omega_2
\end{bmatrix},
\end{equation}
with
\begin{equation}
\mathbf{I} = \begin{bmatrix}
    (I_{zz}^2 - I_{yy}^2) & 0 & 0\\
    0 & (I_{xx}^2 - I_{zz}^2)&0 \\
    0 & 0 & (I_{yy}^2 - I_{xx}^2)
\end{bmatrix}.
\end{equation}
The constant matrix $\mathbf{J}_2$ can be precomputed from the real inertia values. By substituting previous eqs.~(\ref{eqn:w12_bound}) and (\ref{eqn:w3_bound}), we can form the following bound
\begin{equation}\label{eqn:cross_bound}
\mathbf{\Omega} \times \mathbf{J}\mathbf{\Omega} \leq \mathbf{I}\begin{bmatrix}
    \frac{\sqrt{2}m}{2\tau}||\mathbf{j}||\left(|\dot\psi|+
 \frac{m}{\tau}\left|\left|\mathbf{j}\right|\right|\right)\\
     \frac{\sqrt{2}m}{2\tau}||\mathbf{j}||\left(|\dot\psi|+
 \frac{m}{\tau}\left|\left|\mathbf{j}\right|\right|\right)\\
    \frac{m^2}{\tau^2}||\mathbf{j}||^2
\end{bmatrix}.
\end{equation}
We can combine the above inequality $\mathbf{\Omega} \times \mathbf{J}\mathbf{\Omega} \leq K(\dot\psi,\mathbf{a},\mathbf{j},\mathbf{s})$ with eq.~(\ref{eqn:omega_bound}) to create a bound on the moments as
\begin{equation}\label{eqn:moment_bound}
\mathbf{M} \leq \mathbf{J}H\left(\psi,\dot\psi,\ddot\psi,\mathbf{a},\mathbf{j},\mathbf{s}\right)+ K\left(\psi,\dot\psi,\mathbf{a},\mathbf{j},\mathbf{s}\right).
\end{equation}
The relationship between moments and motor speeds is linear which can be solved knowing the vehicle's hyperparameters. This can be applied to motor speed to generate bounds. The above inequalities are combinations of the flat outputs to formulate the bounds on thrust rate, angular velocity, angular acceleration, and moments. Normally, we can also apply this to our global bound checker as additional checks for robots where the low-level motor control setup is not reactive enough. It is also notable that the eqs.~(\ref{eqn:omega_bound}) and~(\ref{eqn:moment_bound}) are functions of the jerk and snap of the polynomial. This implies that if we increase the time and minimize the cost on jerk/snap the following costs will be reduced and will indirectly pull the planned trajectory within the bounds.

\subsubsection{Moment Rate}~\label{sec:moment_rate}
Considering that the moment is linearly related to the quadrotor's motor speed, the moment rate is therefore linearly related to the motor's acceleration. In practice and supported by claims from past works \cite{Mellinger2011}, the quadrotor motor acceleration or moment rates can generally be neglected in planning because motor speed control is almost instantaneous compared to its rigid body motion. For less aggressive platforms, where motor acceleration is a concern, then similar to eq.~(\ref{eqn:moment_bound}) another bound can be established by taking an additional derivative of eq.~(\ref{eqn:moment_det}) and solving again in terms of the flat outputs. As eq.~(\ref{eqn:moment_bound}) is bounded by up to the the fourth order derivative of the position or the quadrotor's angular acceleration, moment rate would be bounded as a function of the flat outputs up to the fifth order derivative or the angular jerk. Several works~\cite{Thomas_perch_inclined,Zhang_optimal_traj,Mellinger2011,ji2022real} do not consider either the fifth order derivative of the flat outputs or the angular jerk in their planning for the aforementioned reasons.

\end{document}